\documentclass{article}

\usepackage{graphicx} 
\usepackage{subfigure}
\usepackage{wrapfig}
\usepackage{xspace}
\usepackage{caption}

\usepackage{algorithm}
\usepackage{algorithmic}

\usepackage{enumitem}
\usepackage{hyperref}

\usepackage{fullpage}
\usepackage{authblk}
\usepackage[round]{natbib} 
\renewcommand{\cite}{\citep}

\usepackage{Definitions}
\usepackage{color}

\renewcommand{\leq}{\leqslant}
\renewcommand{\le}{\leqslant}
\renewcommand{\geq}{\geqslant}
\renewcommand{\ge}{\geqslant}

\title{\huge Learning from Conditional Distributions via \\ Dual Embeddings}

\author{
    Bo Dai$^1$, Niao He$^2$, Yunpeng Pan$^1$, Byron Boots$^1$, Le Song$^1$\\
    $^1$ Georgia Institute of Technology\\
    \{bodai, ypan37\}@gatech.edu, \{lsong, bboots\}@cc.gatech.edu\\
    $^2$ University of Illinois at Urbana-Champaign\\
    niaohe@illinois.edu
}

\begin{document}
\maketitle

\begin{abstract}
  Many machine learning tasks, such as learning with invariance and policy evaluation in reinforcement learning, can be characterized as problems of \emph{learning from conditional distributions}. In such problems, each sample $x$ itself is associated with a conditional distribution $p(z|x)$ represented by samples $\{z_i\}_{i=1}^M$, and the goal is to learn a function $f$ that links these conditional distributions to target values $y$. These learning problems become very challenging when we only have limited samples or in the extreme case only one sample from each conditional distribution. Commonly used approaches either assume that $z$ is independent of $x$, or require an overwhelmingly large samples from each conditional distribution. 

  To address these challenges, we propose a novel approach which employs a new \emph{min-max reformulation} of the learning from conditional distribution problem.  With such new reformulation, we only need to deal with the \emph{joint distribution} $p(z,x)$. We also design an efficient learning algorithm, \emph{Embedding-SGD}, and establish theoretical sample complexity for such problems. Finally, our numerical experiments on both synthetic and real-world datasets show that the proposed approach can significantly improve over the existing algorithms.  
\end{abstract}

\section{Introduction}\label{sec:intro}

\setlength{\abovedisplayskip}{3pt}
\setlength{\abovedisplayshortskip}{3pt}
\setlength{\belowdisplayskip}{3pt}
\setlength{\belowdisplayshortskip}{3pt}
\setlength{\jot}{2pt}

\setlength{\floatsep}{2ex}
\setlength{\textfloatsep}{2ex}

We address the problem of \emph{learning from conditional distributions} where the goal is to learn a function that links conditional distributions to target variables. Specifically, we are provided input samples $\{x_i\}_{i=1}^N\in\Xcal^N$ and their corresponding responses $\{y_i\}_{i=1}^N\in \Ycal^N$. For each $x\in\Xcal$, there is an associated conditional distribution $p(z|x): \Zcal\times \Xcal\rightarrow \RR$. However, we cannot access the entire conditional distributions $\{p(z|x_i)\}_{i=1}^N$ directly; rather, we only observe a limited number of samples or in the extreme case only \emph{one sample} from each conditional distribution $p(z|x)$. The task is to learn a function $f$ which links the conditional distribution $p(z|x)$ to target $y \in \Ycal$ by minimizing the expected loss: 
\begin{equation}\label{eq:target}
\min_{f\in\Fcal}~L(f)=\EE_{x,y}\sbr{\ell\rbr{y, \EE_{z|x}\sbr{f(z,x)}}}
\end{equation}
where $\ell:\Ycal \times\Ycal \to\RR$ is a convex loss function. The function space $\Fcal$ can be very general, but we focus on the case when $\Fcal$ is a reproducing kernel Hilbert space~(RKHS) in main text, namely,
$\Fcal = \{f:\Zcal\times \Xcal\rightarrow \RR \,|\, f(z, x) = \inner{f}{\psi(z, x)}\}$ 
where $\psi(z, x)$ is a suitably chosen (nonlinear) feature map. Please refer to Appendix~\ref{appendix:extend_dual_embedding} for the extension to arbitrary function approximators, \eg, random features and neural networks.

The problem of learning from conditional distributions appears in many different tasks. For example:
\begin{itemize}[leftmargin=*,nosep,nolistsep]
    \item {\bf Learning with invariance.} Incorporating priors on invariance into the learning procedure is crucial for computer vision~\cite{NiyGirPog98}, speech recognition~\cite{AnsLeiRosMut13} and many other applications. The goal of invariance learning is to estimate a function which minimizes the expected risk while at the same time preserving consistency over a group of operations $g=\{g_j\}_{j=1}^\infty$. \citet{MroVoiPog15} shows that this can be accomplished by solving the following optimization problem
    \begin{equation}\label{eq:invariant}
      \min_{f\in \tilde\Hcal} \EE_{x, y} [\ell(y, \EE_{z|x\sim \mu(g(x))}[\langle f, \psi(z)\rangle_{\tilde\Hcal}] )]  +(\nu/2)\|f\|_{\tilde\Hcal}^2
    \end{equation}
    where $\tilde\Hcal$ is the RKHS corresponding to kernel $\tilde k$ with implicit feature map $\psi(\cdot)$, $\nu>0$ is the regularization parameter. Obviously, the above optimization~(\ref{eq:invariant}) is a special case of (\ref{eq:target}). In this case, $z$ stands for possible variation of data $x$ through conditional probability given by some normalized Haar measure $\mu(g(x))$. Due to computation and memory constraints,  one can only afford to generate a few virtual samples from each data point $x$.\\    

    \item {\bf Policy evaluation in reinforcement learning.} Policy evaluation is a fundamental task in reinforcement learning. Given a policy $\pi(a|s)$ which is a distribution over action space condition on current state $s$, the goal is to estimate the value function $V^\pi(\cdot)$ over the state space. $V^\pi(s)$ is the fixed point of the Bellman equation 
    $$
    V^\pi(s) = \EE_{s'|a, s}[R(s,a)+\gamma V^\pi(s')],
    $$ 
    where $R(s,a):\Scal\times \Acal\to \RR$ is a reward function and $\gamma\in(0, 1)$ is the discount factor. Therefore, the value function can be estimated from data by minimizing the mean-square Bellman error~\cite{Baird95,SutMaeSze08}:
    \begin{equation}\label{eq:RL_obj}
      \min_{~V^\pi}~\EE_{s, a}\sbr{\rbr{R(s,a) - \EE_{s'|a, s}\sbr{V^\pi(s) - \gamma V^\pi(s')}}^2}.
    \end{equation}
    Restrict the policy to lie in some RKHS, this optimization is clearly a special case of~\eq{eq:target} by viewing $\rbr{(s, a), R(s, a), s'}$ as $(x, y, z)$ in~\eq{eq:target}. Here, given state $s$ and the the action $a\sim \pi(a|s)$, the successor state $s'$ comes from the transition probability $p(s'|a,s)$. Due to the online nature of MDPs, we usually observe only one successor state $s'$ for each action $a$ given $s$,~\ie, only one sample from the conditional distribution given $s, a$.\\ 

    \item{\bf Optimal control in linearly-solvable MDP.} The optimal control in a certain class of MDP, \ie, linearly-solvable MDP, can be achieved by solving the linear Bellman equation~\cite{Todorov06,Todorov09}
    \begin{equation}\label{eq:linear_optimal_bellman}
    z(s) = \exp\rbr{-R(s)}\EE_{s'|s\sim p(s'|s)}\sbr{z(s')},
    \end{equation}
    where $R(s)$ denotes the immediate cost and $p(s'|s)$ denotes the passive dynamics without control. With $z(s)$, the trajectory of the optimal control $\pi^*$ can be calculated by $p^{\pi^*}(s'|s) = \frac{p(s'|s)z(s)}{\EE_{s'|s\sim p(s'|s)}\sbr{z(s')}}$. Therefore, $z(\cdot)$ can be estimated from data by optimizing
    \begin{equation}\label{eq:OC_obj}
      \min_{~z}~\EE_{s}\sbr{\rbr{z(s) - \EE_{s'|s}\sbr{\exp\rbr{-R(s)}z(s')}}^2}.
    \end{equation}
    Restricting function $z(\cdot)$ to lie in some RKHS, this optimization is a special case of~\eq{eq:target} by viewing $\rbr{s, 0, s'}$ as  $(x, y, z)$ in~\eq{eq:target}. Here given a state $s$, the successor state $s'$ comes from the passive dynamics. Similar as policy evaluation, we usually observe only one successor state $s'$ given $s$,~\ie, only one sample from the conditional distribution given $s$.\\ 

    \item{\bf Hitting time and stationary distribution in stochastic process.} Estimating the hitting time and stationary distribution of stochastic process are both important problems in social network application and MCMC sampling technique. Denote the transition probability as $p(s'|s)$. The hitting time of $s_j$ starting from $s_i$ is defined as $H(s_i, s_j) = \inf\cbr{n\ge 0; S_n = s_j, S_0 = s_i}$. Hence, the expected hitting time $h(\cdot, \cdot)$ satisfies
    \begin{eqnarray}
    h(s_i, s_j) =  \begin{cases}
     1 + \EE_{s_k\sim p(s|s_i), s_k\neq s_j}\sbr{h(s_k, s_j)} & \quad \text{if } i\neq j\\
    1 + \EE_{s_k\sim p(s|s_i)}\sbr{h(s_k, s_i)}  & \quad \text{if } i = j\\
   \end{cases}.
  \end{eqnarray}
  Based on the property of stochastic process, we can obtain the stationary distribution with $\pi(s_i) = \frac{1}{h(s_i, s_i)}$. The hitting time $h(\cdot, \cdot)$ can be learned by minimizing:
  \begin{eqnarray}\label{eq:HT_obj}
  \min_{h}\EE_{s, t}\sbr{\rbr{1 - \EE_{s'\sim \tilde p(s'|s, t)}\sbr{h(s, t) - h(s', t)}}^2},
  \end{eqnarray}
  where $\tilde p(s'|s, t) = p(s'|s)$ if $s=t$, otherwise $\tilde p(s'|s, t) \propto \begin{cases}
    p(s'|s)       & \quad \text{if } s'\neq t\\
    0  & \quad \text{if } s' = t\\
  \end{cases}$. Similarly, when restricting the expected hitting time to lie in some RKHS, this optimization is a special case of~\eq{eq:target} by viewing $\rbr{(s, t), 1, s'}$ as $(x, y, z)$ in~\eq{eq:target}.
  Due to the stochasticity of the process, we only obtain one successor state $s'$ from current state $(s, t)$,~\ie, only one sample from the conditional distribution given $(s, t)$. 
\end{itemize}

\paragraph{Challenges.} Despite the prevalence of learning problems in the form of \eq{eq:target}, solving such problem remains very challenging for two reasons: (\emph{i}) we often have limited samples or in the extreme case only one sample from each conditional distribution $p(z|x)$, making it difficult to accurately estimate the conditional expectation. (\emph{ii}) the conditional expectation is nested inside the loss function, making the problem quite different from the traditional stochastic optimization setting. This type of problem is called {\sl compositional stochastic programming}, and very few results have been established in this domain. 

\paragraph{Related work.} A simple option to address (\ref{eq:target}) is using sample average approximation (SAA), and thus, instead solve 
$$
\min_{f\in\Fcal} \frac{1}{N} \sum_{i=1}^N \sbr{\ell\rbr{y_i, \frac{1}{M}\sum_{j=1}^M f(z_{ij}, x_i)}},
$$ 
where $\{(x_i,y_i)\}_{i=1}^N\sim p(x,y)$, and $\{z_{ij}\}_{j=1}^M\sim p(z|x_i)$ for each $x_i$. To ensure an excess risk of $\epsilon$, both $N$ and $M$ need be at least as large as $\Ocal(1/\epsilon^2)$, making the overall sample required to be $\Ocal(1/\epsilon^4)$; see~\cite{NemJudLanSha09,WanFanLiu14} and references therein. Hence, when $M$ is small, SAA would provide poor results. 

A second option is to resort to stochastic gradient methods (SGD). One can construct a \emph{biased} stochastic estimate of the gradient using
$
  {\nabla}_f L = \nabla \ell(y, \langle f,\tilde{\psi}(x)\rangle) \tilde{\psi}(x),  
$
where $\tilde{\psi}(x)$ is an estimate of  $\EE_{z|x}[\psi(z,x)]$ for any $x$. To ensure convergence, the bias of the stochastic gradient must be small,~\ie, a large amount of samples from the conditional distribution is needed. 

Another commonly used approach is to first represent the conditional distributions as the so-called kernel conditional embedding, and then perform a supervised learning step on the embedded conditional distributions~\cite{SonFukGre13, GruLevBalPatetal12}. This two-step procedure suffers from poor statistical sample complexity and computational cost. The kernel conditional embedding estimation costs $O(N^3)$, where $N$ is number of pair of samples $(x, z)$. To achieve $\epsilon$ error in the conditional kernel embedding estimation, $N$ needs to be $\Ocal(1/\epsilon^4)$\footnote{With appropriate assumptions on joint distribution $p(x, z)$, a better rate can be obtained~\cite{GruLevBalPatetal12}. However, for fair comparison, we did not introduce such extra assumptions.}. 

Recently,~\citet{WanFanLiu14} solved a related but fundamentally distinct problem of the form, 
\begin{equation}\label{eq:WanFanLiu}
\min_{f\in\Fcal}~L(f)=\EE_{y}\sbr{\ell(y, \EE_{z}[f(z)])}
\end{equation}
where $f(z)$ is a smooth function parameterized by some finite-dimensional parameter.  The authors provide an algorithm that combines stochastic gradient descent with moving average estimation for the inner expectation, and achieves an overall $\Ocal(1/\epsilon^{3.5})$ sample complexity for smooth convex loss functions. The algorithm does not require the loss function to be convex, but it cannot directly handle random variable $z$ with \emph{infinite support}. Hence, such an algorithm does not apply to the more general and difficult situation that we consider in this paper. 

\paragraph{Our approach and contribution.} To address the above challenges, we propose a novel approach called \emph{dual kernel embedding}. The key idea is to reformulate (\ref{eq:target}) into a min-max or saddle point problem by utilizing the Fenchel duality of the loss function. We observe that with smooth loss function and continuous conditional distributions, the dual variables form a continuous function of $x$ and $y$. Therefore, we can parameterize it as a function in some RKHS induced by any universal kernel, where the information about the marginal distribution $p(x)$ and conditional distribution $p(z|x)$ can be aggregated via a kernel embedding of the joint distribution $p(x,z)$. Furthermore, we propose an efficient algorithm based on stochastic approximation to solve the resulting saddle point problem over RKHS spaces, and establish finite-sample analysis of the generic learning from conditional distributions problems. 

Compared to previous applicable approaches, an advantage of the proposed method is that it requires only \emph{one sample} from each conditional distribution. Under mild conditions, the overall sample complexity reduces to $\Ocal(1/\epsilon^2)$ in contrast to the $\Ocal(1/\epsilon^4)$ complexity required by SAA or kernel conditional embedding. As a by-product, even in the degenerate case (\ref{eq:WanFanLiu}), this implies an $\Ocal(1/\epsilon^2)$ sample complexity when inner function is linear, which already surpasses the result obtained in \cite{WanFanLiu14} and is known to be unimprovable. Furthermore, our algorithm is generic for the family of problems of learning from conditional distributions, and can be adapted to problems with different loss functions and hypothesis function spaces.

Our proposed method also offers some new insights into several related applications. In reinforcement learning settings, our method provides the first  algorithm that truly minimizes the mean-square Bellman error~(MSBE) with both theoretical guarantees and sample efficiency. We show that the existing gradient-TD2 algorithm by~\citet{SutMaePreBhaetal09,LiuLiuGhaMah15}, is a special case of our algorithm, and the residual gradient algorithm~\cite{Baird95} is derived by optimizing an upper bound of MSBE. In the invariance learning setting, our method also provides a unified view of several existing methods for encoding invariance. Finally, numerical experiments on both synthetic and real-world datasets show that our method can significantly improve over the previous state-of-the-art performances.

\section{Preliminaries}\label{sec:preliminary}

We first introduce our notations on Fenchel duality, kernel and kernel embedding. Let $\Xcal\subset \RR^d$ be some input space and $k:\Xcal\times\Xcal\to\RR$ be a positive definite kernel function. For notation simplicity, we denote the feature map of kernel $k$ or $\tilde k$ as
$$
\phi(x):= k(x,\cdot),\quad \psi(z) : = \tilde k(z, \cdot),
$$
and use $k(x,\cdot)$ and $\phi(x)$, or $\tilde k(z, \cdot)$ and $\psi(z)$ interchangeably. Then $k$ induces a RKHS $\Hcal$, which has the property $h(x) = \langle h,\phi(x)\rangle_{\Hcal}$, $\forall  h\in \Hcal$, where $\langle\cdot,\cdot\rangle_{\Hcal}$ is the inner product and $\|h\|_\Hcal^2:=\langle h,h\rangle_\Hcal$ is the norm in $\Hcal$. We denote all continuous functions on $\Xcal$ as $\Ccal(\Xcal)$ and $\|\cdot\|_\infty$ as the maximum norm. We call $k$ a \emph{universal kernel}  if $\Hcal$ is dense in $\Ccal(\Omega')$ for any compact set $\Omega'\subseteq\Xcal$,~\ie, for any $\epsilon>0$ and $u\in \Ccal(\Omega')$, there exists $h\in\Hcal$, such that $\|u-h\|_\infty\leq \epsilon$. Examples of universal kernel include the Gaussian kernel, $k(x, x') = \exp\rbr{-\frac{\|x-x'\|_2^2}{\sigma^{-2}}}$, Laplacian kernel, $k(x, x') =\exp\rbr{-\frac{\|x-x'\|_1}{\sigma^{-1}}}$, and so on. 

\paragraph{Convex conjugate and Fenchel duality.} Let $\ell : \RR^d\rightarrow \RR$, its convex conjugate function is defined as
$$
\ell^*(u) = \sup_{v\in \RR^d}\{u^\top v - \ell(v)\}.
$$
When $\ell$ is proper, convex and lower semicontinuous for any $u$, its conjugate function is also proper, convex and lower semicontinuous. More improtantly, the $(\ell, \ell^*)$ are dual to each other, \ie, $(\ell^*)^* = \ell$, which is known as Fenchel duality~\cite{HirLem12,RifLip07}. Therefore, we can represent the $\ell$ by its convex conjugate as ,
$$
\ell(v) = \sup_{u\in \RR^d}\{v^\top u - \ell^*(u)\}.
$$
It can be shown that the supremum achieves if $v\in \partial \ell^*(u)$, or equivalently $u\in \partial \ell(v)$.

\paragraph{Function approximation using RKHS.} Let $\Hcal^\delta:=\{h\in\Hcal:\|h\|_{\Hcal}^2\leq \delta\}$ be a bounded ball in the RKHS, and we define the approximation error of the RKHS $\Hcal^\delta$ as the error from approximating continuous functions in $\Ccal(\Xcal)$ by a function $h\in \Hcal^\delta$, \ie,~\cite{Bach14,Barron93}
\begin{equation}
\begin{array}{c}
\Ecal(\delta):=\sup_{u\in\CC(\Xcal)}\inf_{h\in \Hcal^\delta}\|u-h\|_\infty.
\end{array}
\end{equation}
One can immediately see that $\Ecal(\delta)$ decreases as $\delta$ increases and vanishes to zero as $\delta$ goes to infinity. If $\Ccal(\Xcal)$ is restricted to the set of uniformly bounded continuous functions, then $\Ecal(\delta)$ is also bounded. The approximation property,~\ie,~dependence on $\delta$ remains an open question for general RKHS, but has been carefully established for special kernels. For example, with the kernel $k(x,x') = 1/(1+\exp(\inner{x}{x'}))$ induced by the sigmoidal activation function, we have  $\Ecal(\delta)=O(\delta^{-2/(d+1)}\log(\delta))$ for Lipschitz continuous function space $\Ccal(\Xcal)$~\cite{Bach14}.\footnote{The rate is also known to be unimprovable by~\citet{DeVHowMic89}.}

\paragraph{Hilbert space embedding of distributions.} Hilbert space embeddings of distributions~\cite{SmoGreSonSch07} are mappings of distributions into potentially \emph{infinite} dimensional feature spaces,
\begin{align}
  \mu_{x} \, := \, \EE_{x} \sbr{\phi(x)} \, = \, \int_{\Xcal} \phi(x) p(x) dx~:~ \Pcal \mapsto \Hcal \label{eq:embedding}
\end{align}
where the distribution is mapped to its expected feature map,~\ie,~to a point in a feature space. Kernel embedding of distributions has rich representational power. Some feature map can make the mapping injective~\cite{SriGreFukLanetal08}, meaning that if two distributions are different, they are mapped to two distinct points in the feature space. For instance, when $\Xcal\subseteq\RR^d$, the feature spaces of many commonly used kernels, such as the Gaussian RBF kernel, will generate injective embedding. We can also embed the joint distribution $p(x,y)$ over a pair of variables using two kernels $k(x,x) = \inner{\phi(x)}{\phi(x')}_{\Hcal}$ and $\tilde k(z,z') = \inner{\psi(z)}{\psi(z')}_{\Gcal}$ as  
\begin{eqnarray*}
  \Ccal_{zx} \, &:=& \, \EE_{zx} \sbr{\psi(z)\otimes \phi(x)} \\
   &=& \int_{\Zcal\times\Xcal} \psi(z)\otimes \phi(x) p(z,x) dzdx : \Pcal \mapsto \Hcal\otimes\Gcal, 
\end{eqnarray*}
where the joint distribution is mapped to a point in a tensor product feature space. Based on embedding of joint distributions, kernel embedding of conditional distributions can be defined as $\Ucal_{z|x}:=\Ccal_{zx}\Ccal_{xx}^{-1}$ as an operator $\Hcal \mapsto \Gcal$~\cite{SonFukGre13}. With $\Ucal_{z|x}$, we can obtain the expectations easily, \ie,
\begin{eqnarray}
\EE_{z|x}\sbr{g(z)} = \langle g, \langle \Ucal_{z|x}, \phi(x)\rangle_{\Hcal} \rangle_{\Gcal}.
\end{eqnarray}
Both the joint distribution embedding, $\Ccal_{zx}$, and the conditional distribution embedding, $\Ucal_{z|x}$, can be estimated from \iid~samples $\{(x_i, z_i)\}_{i=1}^N$ from $p(x, z)$ or $p(z|x)$, respectively~\cite{SmoGreSonSch07, SonFukGre13}, as
\begin{eqnarray*}
\widehat\Ccal_{zx} = \frac{1}{N}\Psi \Upsilon^\top, ~\text{and}~~~ \widehat \Ucal_{z|x} = \Psi(K + \lambda I)^{-1}\Upsilon^\top,
\end{eqnarray*}
where $\Psi = (\psi(z_1), \ldots, \psi(z_N))$, $\Upsilon = (\phi(x_1), \ldots, \phi(x_N))$, and $K = \Upsilon^\top \Upsilon$. Due to the inverse of $K + \lambda I$, the kernel conditional embedding estimation requires $\Ocal(N^3)$ cost.

\section{Dual Embedding Framework}

In this section, we propose a novel and sample-efficient framework to solve problem~\eq{eq:target}. Our framework leverages Fenchel duality and feature space embedding technique to bypass the difficulties of nested expectation and the need for overwhelmingly large sample from conditional distributions.  We start by introducing the interchangeability principle, which plays a fundamental role in our method. 
\begin{lemma}[interchangeability principle]\label{lem:switch_correct}
Let $\xi$ be a random variable on $\Xi$ and assume  for any $\xi\in \Xi$, function $g(\cdot,\xi):\RR\to(-\infty,+\infty)$ is a proper\footnote{We say $g(\cdot, \xi)$ is proper when $\{u\in 
\RR: g(u, \xi)<\infty\}$ is non-empty and $g(u, \xi)>-\infty$ for $\forall u$.} and upper semicontinuous\footnote{We say $g(\cdot, \xi)$ is upper semicontinuous when $\{u\in \RR: g(u, \xi)<\alpha\}$ is an open set for $\forall \alpha\in\RR$. Similarly, we say $g(\cdot, \xi)$ is lower semicontinuous when $\{u\in \RR: g(u, \xi) >\alpha\}$ is an open set for $\forall\alpha\in \RR$.} concave function. Then
\begin{equation*}
\EE_{\xi}[\max_{u\in\RR}g(u,\xi)] =\max_{u(\cdot)\in \Gcal(\Xi)}\EE_{\xi}[g(u(\xi),\xi)]. 
\end{equation*}
where $\Gcal(\Xi)=\{u(\cdot):\Xi\to\RR\}$ is the entire space of functions defined on support $\Xi$.  
\end{lemma}
The result implies that one can replace the expected value of point-wise optima by the optimum value over a function space. For the proof of lemma~\ref{lem:switch_correct}, please refer to Appendix~\ref{appendix:dualcontinuity}. More general results of interchange between maximization and integration can be found in \cite[Chapter~14]{RocWet98} and \cite[Chapter~7]{ShaDen14}.

\subsection{Saddle Point Reformulation}

Let the loss function $\ell_y(\cdot):=\ell(y,\cdot)$ in \eq{eq:target} be a proper, convex and lower semicontinuous for any $y$. We denote $\ell_y^*(\cdot)$ as the convex conjugate; hence $\ell_y(v) = \max_{u}\{uv-\ell^*_y(u)\}$, which is also a proper, convex and lower semicontinuous function. Using the Fenchel duality, we can reformulate problem~\eq{eq:target} as
\begin{eqnarray}\label{eq:dual_opt}
\min_{f\in\Fcal} \EE_{xy}\bigg[\max_{u\in \RR}  \Big[\EE_{z|x}[f(z,x)]\cdot u - \ell_y^*(u) \Big]\bigg],
\end{eqnarray}
Note that by the concavity and upper-semicontinuity of $-\ell_y^*(\cdot)$, for any given pair $(x,y)$, the corresponding maximizer of the inner function always exists. Based on the  interchangeability principle stated in Lemma~\ref{lem:switch_correct}, we can further rewrite (\ref{eq:dual_opt}) as
\begin{equation}\label{eq:dual_opt_exchange}
\min_{f\in\Fcal} \max_{u(\cdot)\in \Gcal(\Xi)}\Phi(f,u):= \EE_{zx}[f(z,x)\cdot u(x,y)] - \EE_{xy}[\ell_y^*(u(x,y))], 
\end{equation}
where $\Xi=\Xcal\times\Ycal$ and $\Gcal(\Xi)=\{u(\cdot):\Xi\to\RR\}$ is the entire function space on $\Xi$. We emphasize that the $\max$-operator in~\eq{eq:dual_opt} and~\eq{eq:dual_opt_exchange} have different meanings: the one in~\eq{eq:dual_opt} is taking over a single variable, while the other one in~\eq{eq:dual_opt_exchange} is over all possible function $u(\cdot)\in\Gcal(\Xi)$. 

Now that we have eliminated the nested expectation in the problem of interest, and converted it into a stochastic saddle point problem with an additional dual function space to optimize over. By definition, $\Phi(f,u)$ is always concave in $u$ for any fixed $f$. Since $f(z,x) = \inner{f}{\psi(z,x)}$, $\Phi(f,u)$ is also convex in $f$ for any fixed $u$. Our reformulation (\ref{eq:dual_opt_exchange}) is indeed a convex-concave saddle point problem. 

\begin{figure*}[!t]
  \centering
  \begin{tabular}{ccc}
    \includegraphics[width=0.315\textwidth, trim=1cm 0.8cm 1.6cm 1.6cm, clip]{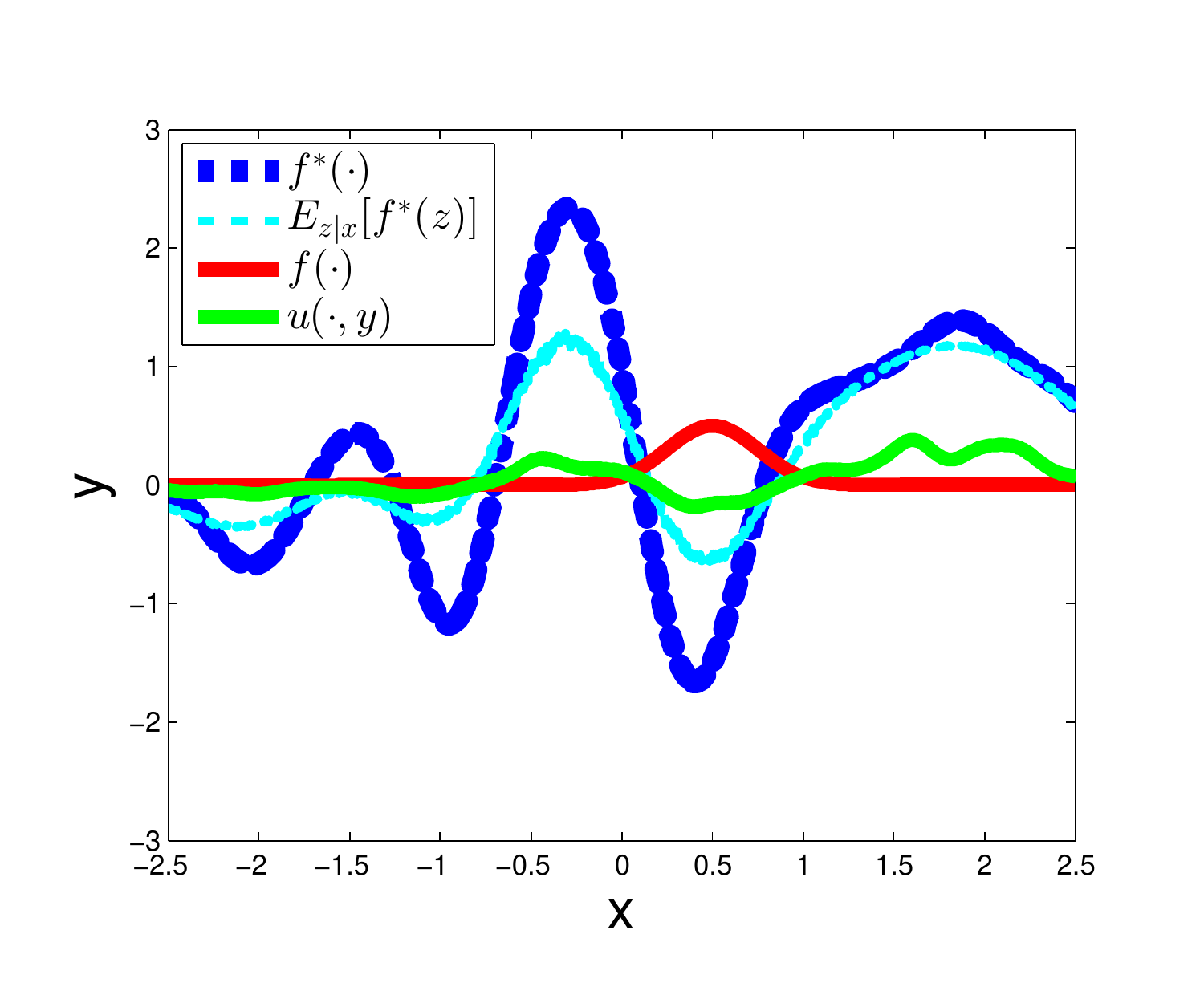}&
    \includegraphics[width=0.315\textwidth, trim=1cm 0.8cm 1.6cm 1.6cm, clip]{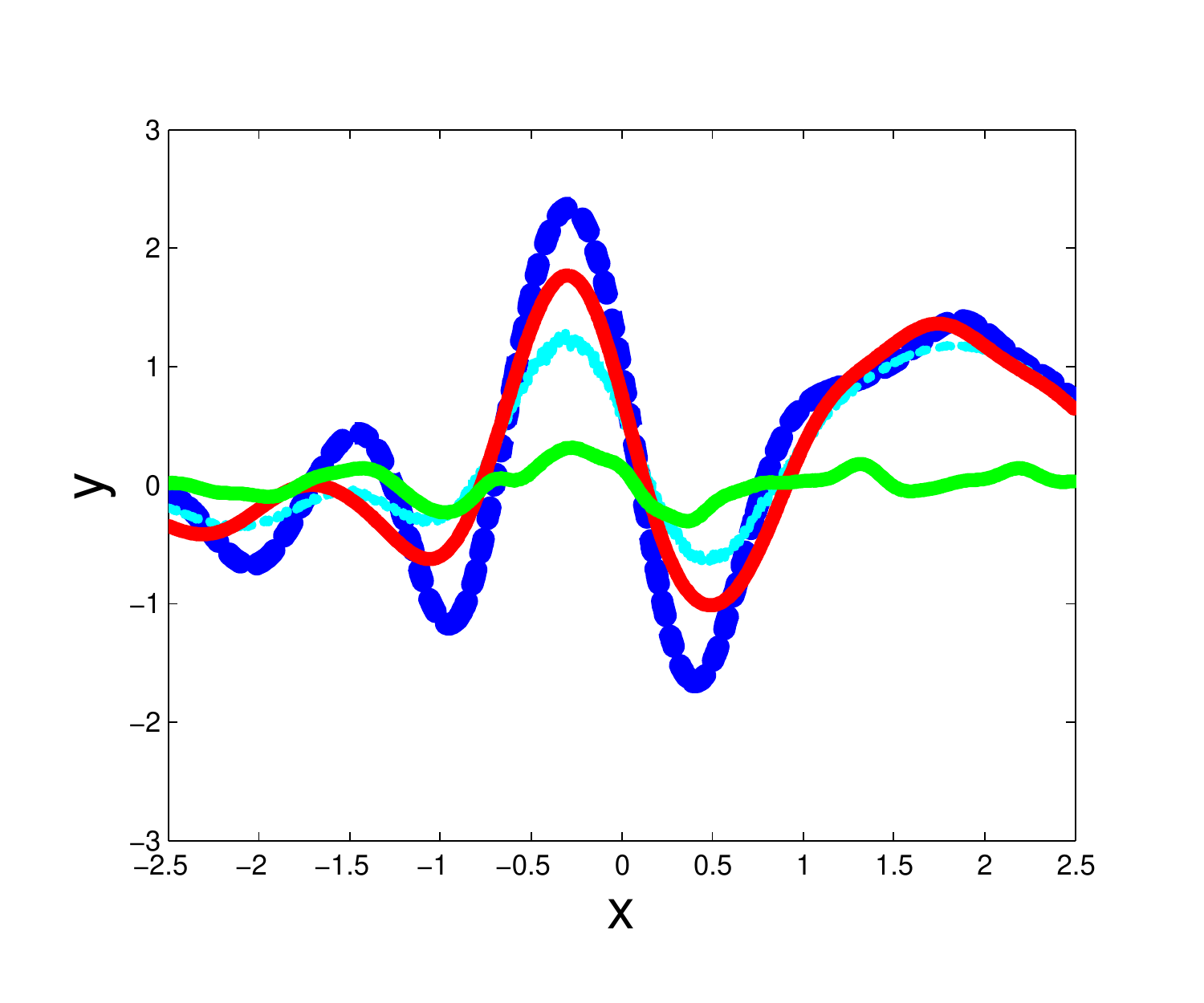} &
    \includegraphics[width=0.315\textwidth, trim=1cm 0.8cm 1.6cm 1.6cm, clip]{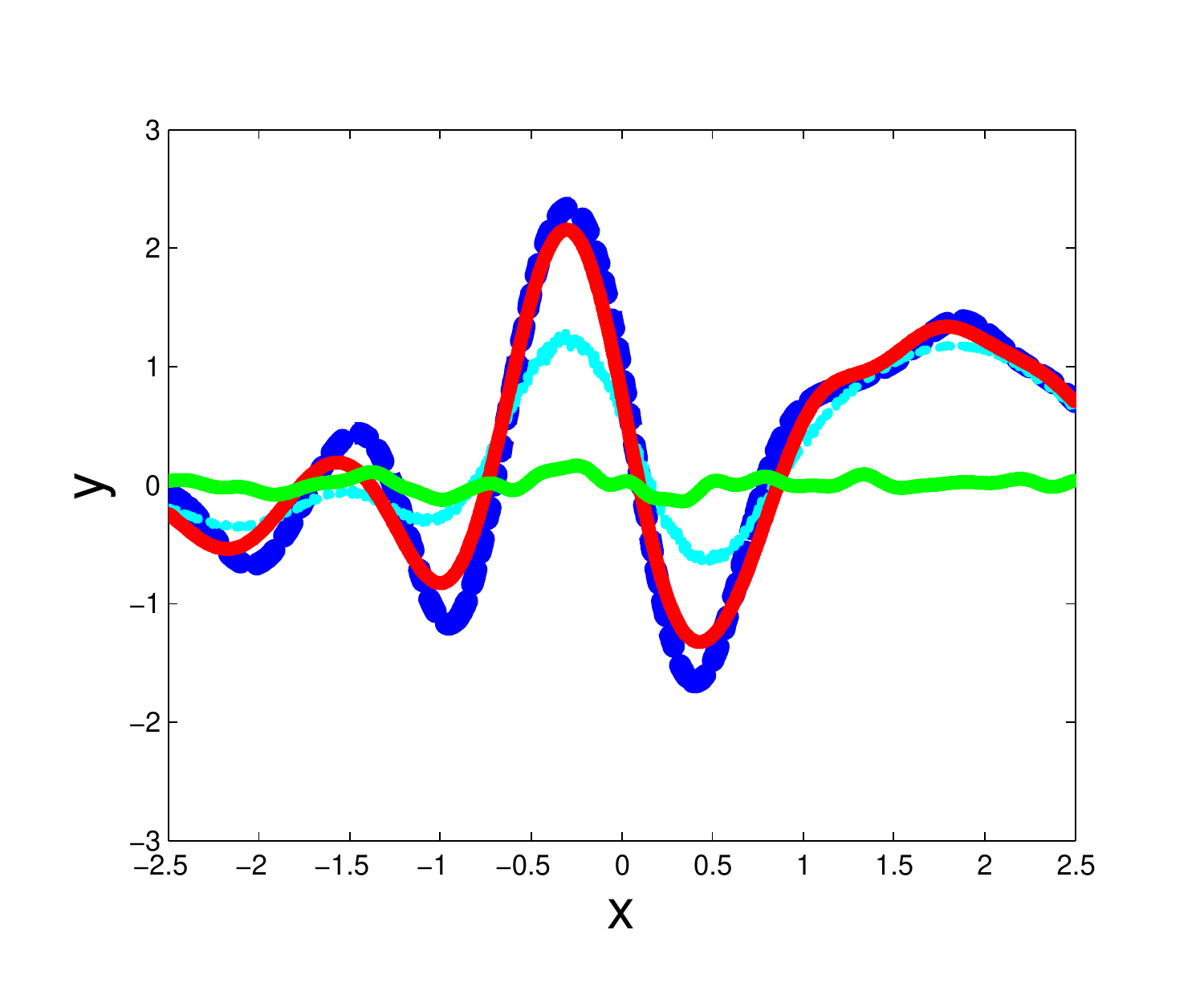} \\
    (a) $0$-th Iteration &(b) $50$-th Iteration &(c) $150$-th Iteration \\

    \includegraphics[width=0.315\textwidth, trim=1cm 0.8cm 1.6cm 1.6cm, clip]{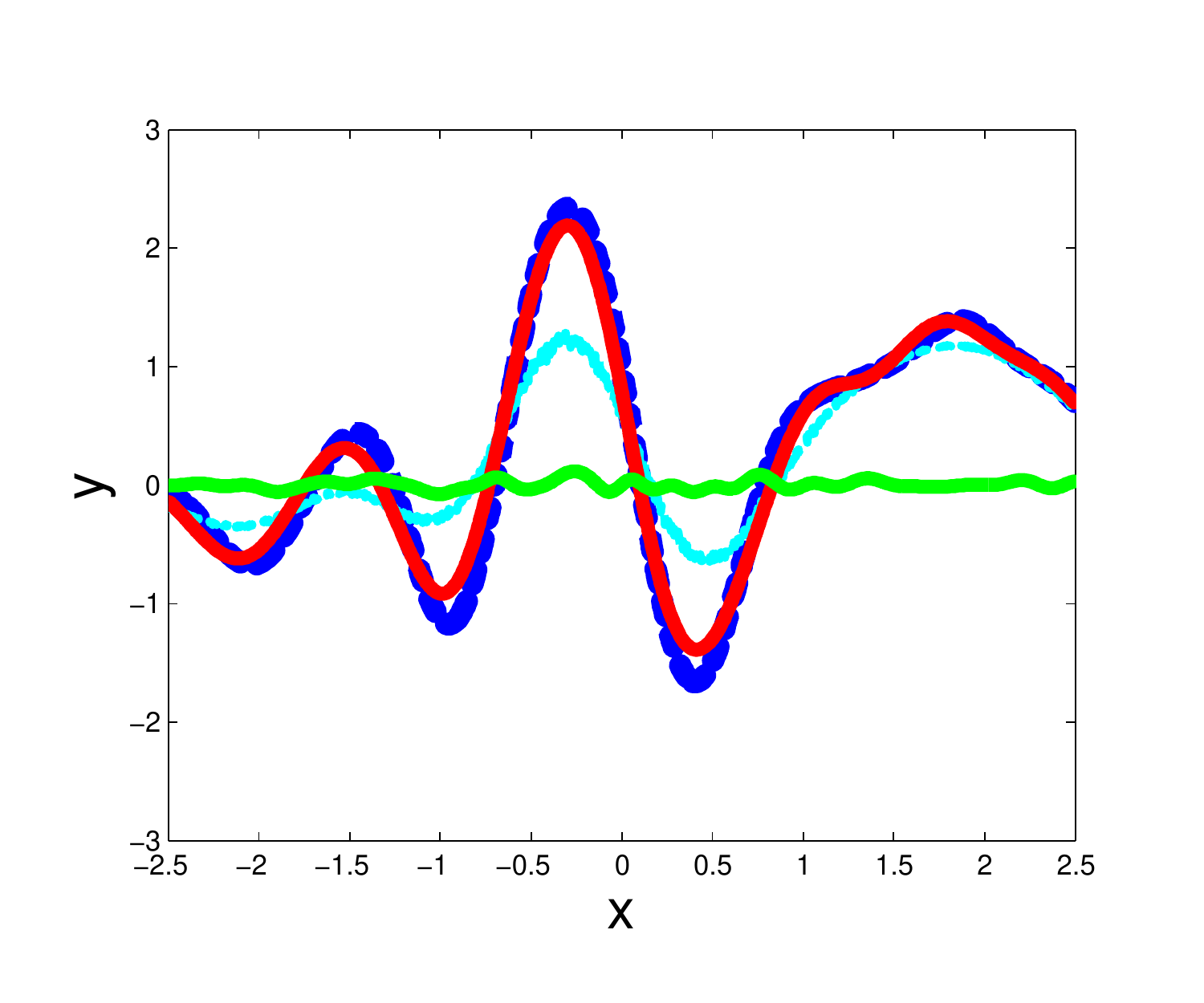} &
    \includegraphics[width=0.315\textwidth, trim=1cm 0.8cm 1.6cm 1.6cm, clip]{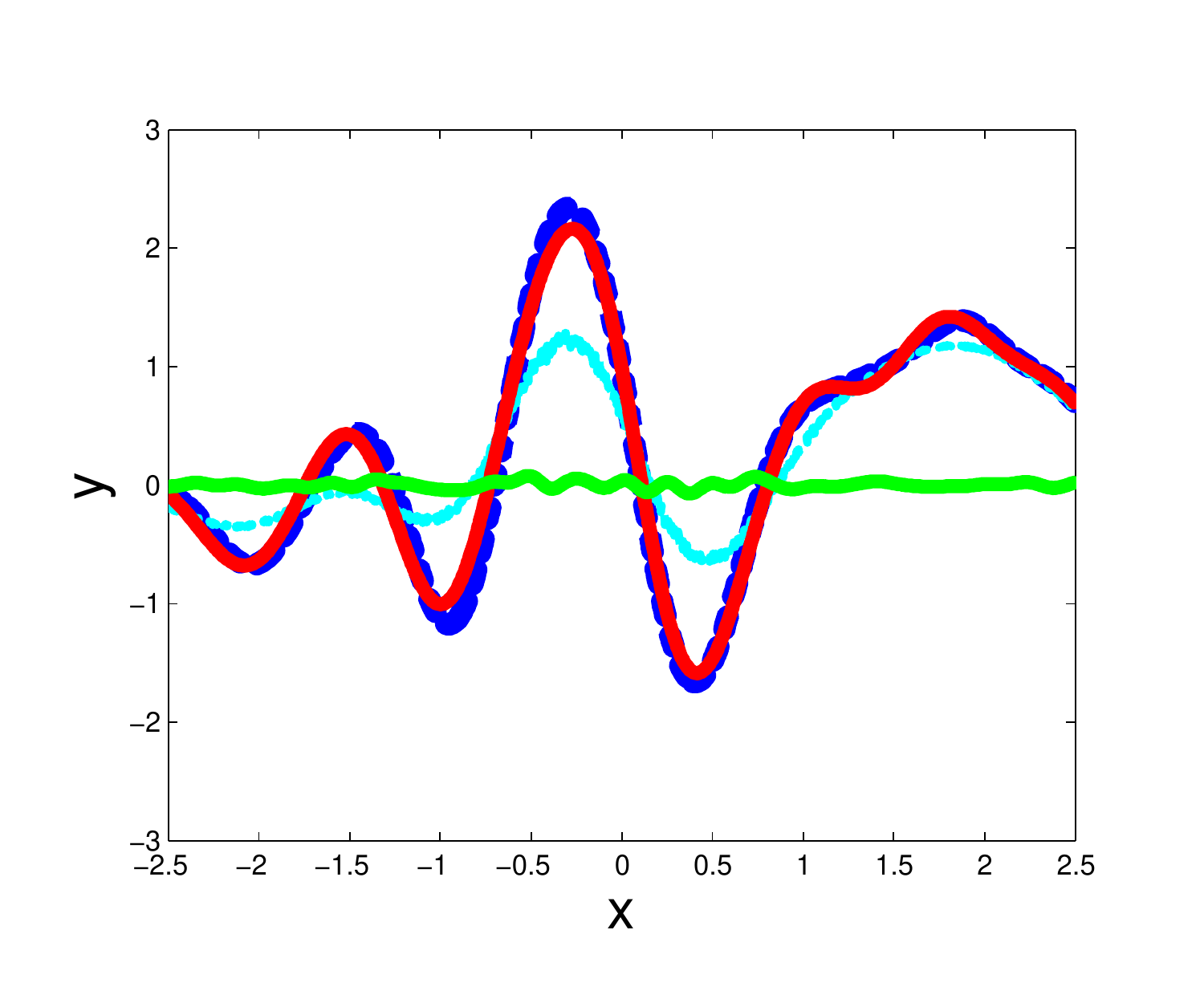} &
    \includegraphics[width=0.315\textwidth, trim=1cm 0.8cm 1.6cm 1.6cm, clip]{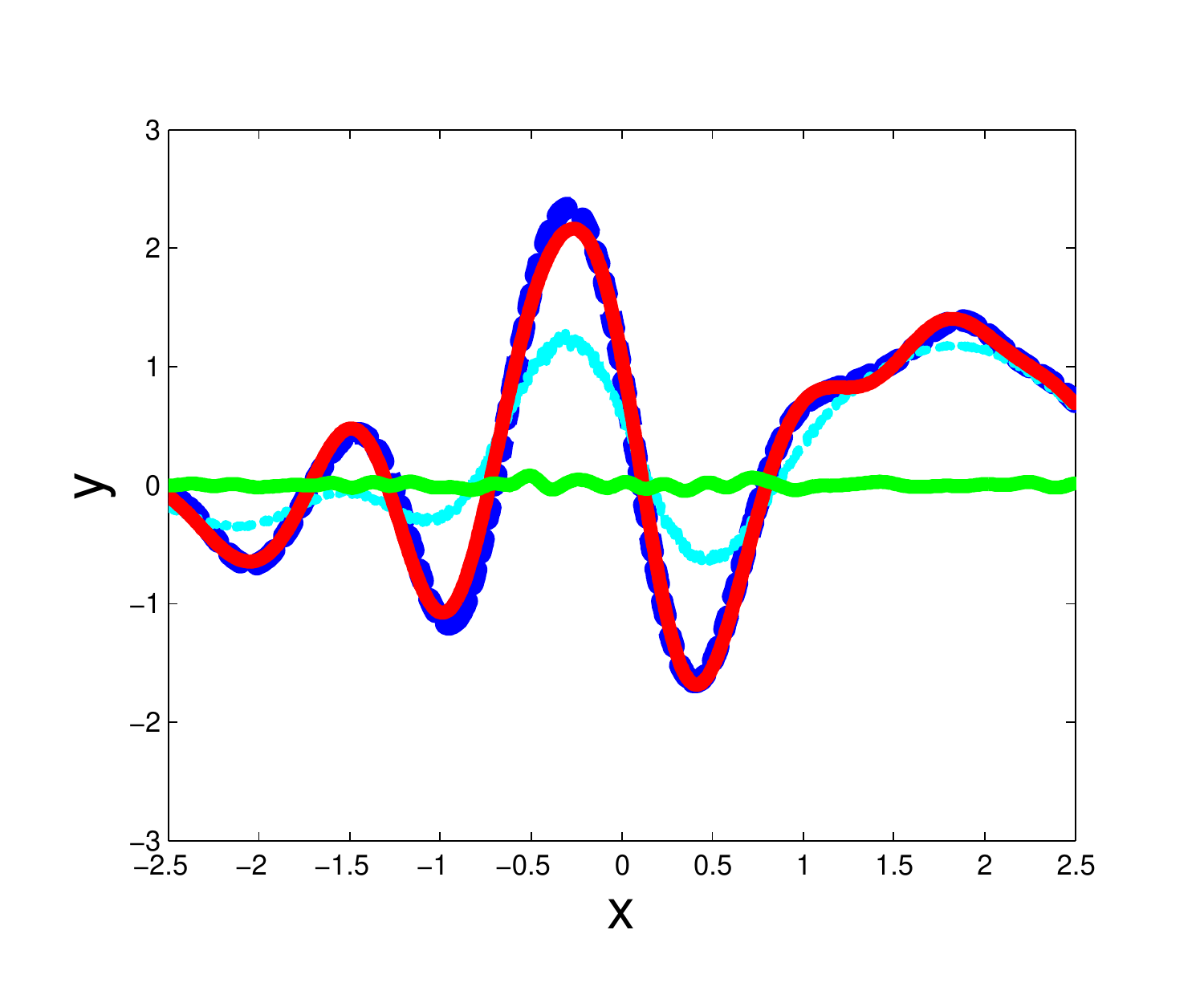} \\
    (a) $400$-th Iteration &(b) $1000$-th Iteration &(c) $2000$-th Iteration \\
  \end{tabular}
  \caption{Toy example with $f^*$ sampled from a Gaussian processes. The $y$ at position $x$ is obtained by smoothing $f^*$ with a Gaussian distribution condition on location $x$, \ie, $y = \EE_{z|x}\sbr{f^*(z)}$ where $z\sim p(z|x) = \Ncal\rbr{x, 0.3}$. Given samples $\{x, y\}$, the task is to recover $f^*(\cdot)$. The blue dash curve is the ground-truth $f^*(\cdot)$. The cyan curve is the observed noisy $y$. The red curve is the recovered signal $f(\cdot)$ and the green curve denotes the dual function $u(\cdot, y)$ with the observed $y$ plugged for each corresponding position $x$. Indeed, the dual function $u(\cdot, y)$ emphasizes the difference between $y$ and $\EE_{z|x}\sbr{f(z)}$ on every $x$. The interaction between primal $f(\cdot)$ and dual $u(\cdot, y)$ results in the recovery of the denoised signal.}
  \label{fig:procedure_illustration}
\end{figure*}

\paragraph{An example.} Let us illustrate this through a concrete example. Let $f^*(\cdot)\in\Fcal$ be the true function, and output $y = \EE_{z|x}\sbr{f^*(z)}$ given $x$. We can recover the true function $f^*(\cdot)$ by solving the optimization problem 
$$
\min_{f\in \Fcal}\EE_{xy}\sbr{\frac{1}{2}\rbr{y - \EE_{z|x}\sbr{f(z)}}^2}.
$$
In this example, $\ell_y(v)=\frac{1}{2}(y-v)^2$ and $\ell_y^*(u)=uy+\frac{1}{2}u^2$. 
Invoking the saddle point reformulation, this leads to 
$$
\min_{f\in \Fcal}\max_{ u\in \Gcal(\Xi)}\EE_{xyz}\sbr{\rbr{f(z) - y}u(x,y)} - \frac{1}{2}\EE_{xy}\sbr{u(x,y)^2}
$$ 
where the dual function $u(x,y)$ fits the discrepancy between $y$ and $\EE_{z|x}\sbr{f(z)}$, and thus, promotes the performance of primal function by emphasizing the different positions. See Figure~\ref{fig:procedure_illustration} for the illustration of the interaction between the primal and dual functions.

\subsection{Dual Continuation}

Although the reformulation in \eq{eq:dual_opt_exchange} gives us more structure of the problem, it is not yet tractable in general. This is because the dual function $u(\cdot)$ can be an arbitrary function which we do not know how to represent. In the following, we will introduce a tractable representation for (\ref{eq:dual_opt_exchange}). 

First, we will define the function $u^*(\cdot):\Xi=\Xcal\times\Ycal\to\RR$ as the \emph{optimal dual function} if for any pair $(x,y)\in\Xi$, 
$$u^*(x,y)\in\argmax\nolimits_{u\in\RR} \left\{u\cdot \EE_{z|x}[f(z,x)] - \ell_y^*(u)\right\}.$$
Note the optimal dual function is well-defined since the optimal set is nonempty. Furthermore, $u^*(x,y)$ is related to the conditional distribution via $u^*(x,y)\in \partial \ell_y(\EE_{z|x}[f(z,x)])$. This can be simply derived from convexity of loss function and Fenchel's inequality; see~\cite{HirLem12} for a more formal argument. Depending on the property of the loss function $\ell_y(v)$, we can further derive that (see proofs in Appendix~\ref{appendix:dualcontinuity}): 
\begin{proposition}\label{prop:dualcontinuity} Suppose both  $f(z,x)$ and $p(z|x)$ are continuous in $x$ for any $z$,
\begin{enumerate}
\item[(1)] (Discrete case) If the loss function $\ell_y(v)$ is continuously differentiable in $v$ for any $y\in\Ycal$, then $u^*(x,y)$ is unique and continuous in $x$  for any $y\in \Ycal$;
\item[(2)] (Continuous case) If the loss function $\ell_y(v)$ is continuously differentiable in $(v,y)$, then $u^*(x,y)$ is unique and continuous in $(x,y)$  on $\Xcal\times\Ycal$.
\end{enumerate}
\end{proposition}
This assumption is satisfied widely in real-world applications. For instance, when it comes to the policy evaluation problem in~\ref{eq:RL_obj}, the corresponding optimal dual function is continuous as long as the reward function is continuous, which is true for many reinforcement tasks. 

The fact that the optimal dual function is a continuous function has interesting consequences. As we mentioned earlier, the space of dual functions can be arbitrary and difficult to represent. Now we can simply restrict the parametrization to the space of continuous functions, which is tractable and still contains the global optimum of the optimization problem in~\eq{eq:dual_opt_exchange}. This also provides us the basis for using an RKHS to approximate these dual functions, and simply optimizing over the RKHS.

\subsection{Feature Space Embedding}

In the rest of the paper, we assume conditions described in Proposition~\ref{prop:dualcontinuity} always hold. For the sake of simplicity, we focus only on the case when $\Ycal$ is a continuous set. Hence, from Proposition~\ref{prop:dualcontinuity}, the optimal dual function is indeed continuous in $(x,y)\in\Xi=\Xcal\times\Ycal$. As an immediate consequence, we lose nothing by restricting the dual function space $\Gcal(\Xi)$ to be continuous function space on $\Xi$. Recall that with the universal kernel, we can approximate any continuous function with arbitrarily small error.  Thus we approximate the dual space $\Gcal(\Xi)$ by the bounded RKHS $\Hcal^{\delta}$ induced by a universal kernel $k((x,y), (x',y')) = \langle \phi(x,y), \phi(x',y')\rangle_{\Hcal}$ where $\phi(\cdot)$ is the implicit feature map. Therefore, $u(x,y) = \inner{u}{\phi(x,y)}_{\Hcal}$. Note that $\Hcal^\delta$ is a subspace of the continuous function space, and hence is a subspace of the dual space $\Gcal(\Xi)$. To distinguish inner 
product between the primal function space $\Fcal$ and the dual RKHS $\Hcal^\delta$, we denote the inner product in $\Fcal$ as $\langle \cdot, \cdot\rangle_{\Fcal}$.

We can rewrite the saddle point problem in~\eq{eq:dual_opt_exchange} as
\begin{align}\label{eq:dual_approximate}
\min_{f \in\Fcal} \max_{u\in \Hcal^\delta} \Phi(f,u) &= \EE_{xyz}\sbr{\langle f,\psi(z,x)\rangle_{\Fcal}\cdot\langle u, \phi(x,y)\rangle_{\Hcal} \hspace{-1mm}-\hspace{-1mm} \ell_y^*(\langle u, \phi(x,y)\rangle_{\Hcal})} \nonumber\\
 &= f^\top \Ccal_{zxy} u - \EE_{xy}[\ell_y^*(\langle u, \phi(x,y)\rangle_{\Hcal})], 
\end{align}
where $f(z, x) = \langle f, \psi(z, x)\rangle_{\Fcal}$ by the definition of $\Fcal$, and $\Ccal_{zxy} = \EE_{zxy}[\psi(z,x)\otimes\phi(x,y)]$ is the joint embedding of $p(z,x, y)$ over $\Fcal\times \Hcal$. The new saddle point approximation (\ref{eq:dual_approximate}) based on dual kernel embedding allows us to efficient represent the dual function and get away from the fundamental difficulty with insufficient sampling from the conditional distribution. There is no need to access either the conditional distribution $p(z|x)$, the conditional expectation $\EE_{z|x}\sbr{\cdot}$, or the conditional embedding operator $\Ucal_{z|x}$ anymore, therefore, reducing both the statistical and computational complexity. 

Specifically, given a pair of sample $(x,y,z)$, where $(x,y)\sim p(x,y)$ and $z\sim p(z|x)$, we can now easily construct an unbiased stochastic estimate for the gradient, namely,
\begin{eqnarray*} 
\nabla_{{f}}\hat{\Phi}_{x,y,z}({f},u)&=&\psi(z,x) u(x,y),\\
\nabla_{u}\hat{\Phi}_{x,y,z}({f},u)&=&[f(z,x)-\nabla \ell_y^*(u(x,y))] \phi(x,y),
\end{eqnarray*}
with $\EE\sbr{\nabla\hat{\Phi}_{x,y,z}({f},u)}=\nabla\Phi(f, u)$, respectively. For simplicity of notation, we use $\nabla$ to denote the subgradient as well as the gradient. With the unbiased stochastic gradient, we are now able to solve the approximation problem (\ref{eq:dual_approximate}) by resorting to the powerful mirror descent stochastic approximation framework~\cite{NemJudLanSha09}.

\subsection{Sample-Efficient Algorithm}

The algorithm is summarized in Algorithm~\ref{alg:stochastic_composite_general}. At each iteration,  the algorithm performs a projected gradient step  both for  the primal variable $f$ and dual variable $u$ based on the unbiased stochastic gradient. The proposed algorithm avoids the need for overwhelmingly large sample sizes from the conditional distributions when estimating the gradient. At each iteration, only one sample from the conditional distribution is required in our algorithm!  

Throughout our discussion, we make the following standard assumptions:
\begin{assumption}\label{asp:bounded}
There exists constant scalars $C_\Fcal$, $M_\Fcal$, and $c_\ell$, such that for any ${f}\in{\Fcal},u\in\Hcal^\delta$,
\begin{eqnarray*}
\EE_{z,x}[\|f(z,x)\|_2^2]\leq M_\Fcal, \quad\EE_{z,x}[\|\psi(z,x)\|_\Fcal^2]\leq C_\Fcal, \quad \EE_{y}[\|\nabla \ell^*_y(u)\|_2^2]\leq c_\ell.
\end{eqnarray*}
\end{assumption}
\begin{assumption}\label{asp:kernel}
 There exists constant $\kappa>0$ such that $k(w,w')\leq \kappa$ for any $w,w'\in\Xcal$.
\end{assumption}
Assumption~\ref{asp:bounded} and~\ref{asp:kernel} basically suggest that the variance of our stochastic gradient estimate is always bounded. Note that we do not assume any strongly convexity or concavity of the saddle point problem, or Lipschitz smoothness. Hence, we set the output as the average of  intermediate solutions weighted by the learning rates $\{\gamma_i\}$, as often used in the literature, to ensure the convergence of the algorithm. 
\begin{algorithm}[t!]
\caption{\textbf{Embedding-SGD} for Optimization~(\ref{eq:dual_approximate})}
  \text{\bf Input:} $p(x,y),\, p(z|x),\, \psi(z,x),\,\phi(x,y),\, \{\gamma_i\geq 0\}_{i=1}^t$\\[-4mm]
  \begin{algorithmic}[1]\label{alg:stochastic_composite_general}
    \FOR{$i=1,\ldots, t$}
      \STATE Sample $(x_i,y_i) \sim p(x,y)$ and $z_i \sim p(z|x)$.      
      \STATE ${f}_{i+1} = \Pi_{\Fcal}({f}_i-\gamma_i\psi(z_i,x_i)u_i(x_i,y_i))$.
      \STATE $u_{i+1} = \Pi_{\Hcal^\delta}(u_i+\gamma_i [f_i(z_i,x_i)-\nabla \ell_{y_i}^*(u_i(x_i,y_i))]\phi(x_i,y_i))$
    \ENDFOR\\
    \text{\bf Output:}  $\bar{f}_t=\frac{\sum_{i=1}^t\gamma_i{f}_i}{\sum_{i=1}^t\gamma_i}$, $\bar u_t=\frac{\sum_{i=1}^t \gamma_iu_i}{\sum_{i=1}^t \gamma_i}$
      \end{algorithmic}
\end{algorithm}

Define the accuracy of any candidate solution $(\bar f,\bar u)$ to the saddle point problem as
\begin{equation}\label{eq:sadaccuracy}
\epsilon_{\rm gap}(\bar f,\bar u):=\max_{u\in \Hcal^\delta}\Phi(\bar f,u)-\min_{f\in\Fcal}\Phi(f,\bar u).
\end{equation}
We have the following convergence result, 
\begin{theorem}
\label{thm:main}
 Under Assumptions~\ref{asp:bounded} and~\ref{asp:kernel}, the solution $(\bar{f}_t,\bar u_t)$ after $t$ steps of the algorithm with step-sizes being $\gamma_t=\frac{\gamma}{\sqrt{t}} (\gamma>0)$ satisfies:
\begin{equation}\label{eq:mainbound}
\EE[\epsilon_{\rm gap} (\bar{f}_t,\bar  u_t)]\leq [(2D_{\Fcal}^2+4\delta)/\gamma+\gamma \Ccal(\delta,\kappa) ]\frac{1}{\sqrt{t}}
\end{equation}
where $D_{\Fcal}^2=\sup_{{f}\in{\Fcal}}\frac{1}{2}\|{f}_0-{f}\|_2^2$  and $\Ccal(\delta,\kappa)= \kappa(5M_\Fcal+c_\ell)+\frac{1}{8}(\delta+\kappa)^2C_\Fcal$. 
\end{theorem}
The above theorem implies that our algorithm achieves an overall $\Ocal(1/\sqrt{t})$ convergence rate, which is known to be unimprovable already for traditional stochastic optimization with general convex loss function \cite{NemJudLanSha09}. We further observe that 
\begin{proposition}
\label{prop:Lipschitzianity}
If $f(z,x)$ is uniformly bounded by $C$ and $\ell_y^*(v)$ is uniformly $K$-Lipschitz continuous in $v$ for any $y$, then  $\Phi(f,u)$ is $(C+K)$-Lipschitz continuous on $\Gcal(\Xi)$ with respect to $\|\cdot\|_\infty$, i.e. 
$$|\Phi(f,u_1)-\Phi(f,u_2)|\leq (C+K)\|u_1-u_2\|_\infty, \forall u_1,u_2\in\Gcal(\Xi).$$
\end{proposition}

Let ${f}_*$ be the optimal solution to (\ref{eq:target}). Invoking the Lipschitz continuity of $\Phi$ and using standard arguments of decomposing the objective, we have
$$L(\bar {f}_t)-L({f}_*)\leq  \epsilon_{\rm gap}(\bar {f}_t,\bar u_t)+2(C+K)\Ecal(\delta).$$
Combining Proposition~\ref{prop:Lipschitzianity} and Theorem~\ref{thm:main}, we finally conclude that under the conditions therein,  
\begin{equation}\label{eq:combinedbound}
\EE[L(\bar{f}_t)-L({f}_*)]\leq \Ocal\left(\frac{\delta^{3/2}}{\sqrt{t}} +\Ecal(\delta)\right).
\end{equation}
There is clearly a delicate trade-off between the optimization error and approximation error. Using large $\delta$  will increase the optimization error but decrease the approximation error. When $\delta$ is moderately large (which is expected in the situation when the optimal dual function has small magnitude), our dual kernel embedding algorithm can achieve an overall $\Ocal(1/\epsilon^2)$ sample complexity when solving learning problems in the form of (\ref{eq:target}). For the analysis details, please refer to Appendix~\ref{appendix:convergece_rate}.

\section{Applications}\label{sec:application}

In this section, we discuss in details how the dual kernel embedding can be applied to solve several important learning problems in machine learning, \eg, learning with invariance and reinforcement learning, which are the special cases of the optimization~\eq{eq:target}. {By simple verification, these examples satisify our assumptions for the convergence of algorithm.} We tailor the proposed algorithm for the respective learning scenarios and unify several existing algorithms for each learning problem into our framework. Due to the space limit, we only focus on algorithms with kernel embedding. Extended algorithms with random feature, doubly SGD, neural networks as well as their hybrids can be found in Appendix~\ref{appendix:dual_random_fea},~\ref{appendix:doublySGD} and~\ref{appendix:dual_neural_networks}.

\subsection{Learning with Invariant Representations} 

{\bf Invariance learning.} The goal is to solve the optimization (\ref{eq:invariant}), which learns a function in RKHS $\tilde\Hcal$ with kernel $\tilde k$. Applying the dual kernel embedding, we end up solving the saddle point problem
\begin{eqnarray*}
\min_{f\in\tilde{\Hcal}}\max_{u\in \Hcal}~\EE_{zxy}\sbr{\langle f, \psi(z)\rangle_{\tilde\Hcal} \cdot u(x,y)}- \EE_{xy}[\ell_y^*(u(x,y))]+ \frac{\nu}{2}\|f\|^2_{\tilde{\Hcal}}, 
\end{eqnarray*}
where $\Hcal$ is the dual RKHS with the universal kernel introduced in our method. 

\noindent{\bf Remark.} The proposed algorithm bears some similarities to virtual sample techniques~\cite{NiyGirPog98, LooCanBot07} in the sense that they both create examples with prior knowledge to incorporate invariance. In fact, the virtual sample technique can be viewed as optimizing an upper bound of the objective~\eq{eq:invariant} by simply moving the conditional expectation outside, \ie,
$\EE_{x, y} [\ell(y, \EE_{z|x}[f(z)] )] \le \EE_{x, y, z}\big[\ell(y,f(z))\big]$,
where the inequality comes from convexity of $\ell(y, \cdot)$.

\noindent{\bf Remark.} The learning problem~\eq{eq:invariant} can be understood as learning with RKHS $\hat \Hcal$ with Haar-Integral kernel $\hat k$ which is generated by $\tilde k$ as $\hat k(x, x') =\langle \EE_{p(z|x)}[\psi(z)], \EE_{p(z'|x')}[\psi(z')]\rangle_{\tilde \Hcal}$, with implicit feature map $\EE_{p(z|x)}[\psi(z)]$. If $f\in \tilde \Hcal$, then, $ f(x) = \EE_{z|x}[\langle f, \psi(z)\rangle_{\tilde\Hcal}] = \langle f, \EE_{z|x}[\psi(z)]\rangle \in \hat \Hcal$.
The Haar-Integral kernel can be viewed as a special case of Hilbertian metric on probability measures on which the output of function should be invariant~\cite{HeiBou05}. Therefore, other kernels defined for distributions, \eg, the probability product kernel~\cite{JebKonHow04}, can be also used in incorporating invariance.

\noindent{\bf Remark.} Robust learning with contamined samples can also be viewed as incorporating invariance prior with respect to the perturbation distribution into learning procedure. Therefore, rather than resorting to robust optimization techniques~\cite{BhaPanSmo05,BenGhaNem08}, the proposed algorithm for learning with invariance serves as a viable alternative for robust learning.

\subsection{Reinforcement Learning} 

{\bf Policy evaluation.} The goal is to estimate the value function $V^\pi(\cdot)$ of a given policy $\pi(a|s)$ by minimizing the mean-square Bellman error (MSBE)~\eq{eq:RL_obj}. With $V^\pi\in \tilde\Hcal$ with feature map $\psi(\cdot)$, we apply the dual kernel embedding, which will lead to the saddle point problem 
\begin{eqnarray}\label{eq:RL_dual}
\min_{V^\pi\in\tilde{\Hcal}} \max_{u\in\Hcal} &&\EE_{s',a, s} \sbr{\rbr{R(s,a) - \langle V^\pi,  \psi(s) -\gamma \psi(s') \rangle_{\tilde \Hcal}} u(s, a)} -\frac{1}{2}\EE_{s, a}[u^2(s, a)].
\end{eqnarray}
In the optimization~\eq{eq:RL_dual}, we simplify the dual $u$ to be function over $\Scal \times \Acal$ due to the fact that $R(s, a)$ is determinastic given $s$ and $a$ in our setting. If $R(s, a)$ is a random variable sampled from some distribution $p(R|s, a)$, then the dual function should be defined over $\Scal \times \Acal \times \RR$.

\noindent{\bf Remark.} The algorithm can be extended to off-policy setting. Let $\pi_b$ be the behavior policy and $\rho(a|s) = \frac{\pi(a|s)}{\pi_b(a|s)}$ be the importance weight, then the objctive will be adjusted by $\rho(a|s)$, \ie 
\begin{eqnarray*}
\min_{V^\pi\in \tilde \Hcal}\EE_{s, a}\sbr{\rbr{\EE_{s'|a, s}\sbr{\rho(a|s)\rbr{R(s, a) - \langle V^\pi, \psi(s) -\gamma \psi(s')\rangle_{\tilde \Hcal} }}}^2}
\end{eqnarray*}
where the successor state $s'\sim P(s'|s, a)$ and actions $a\sim\pi_b(a|s)$ from behavior policy. With the dual kernel embedding, we can derive similar algorithm for off-policy setting, with extra importance weight $\rho(a|s)$ to adjust the sample distribution.

\noindent{\bf Remark.} We used different RKHSs for primal and dual functions. If we use the \emph{same finite basis functions} to parametrize both the value function and the dual function, \ie, $V^\pi(s)=\theta^T\psi(s)$ and $u(s)=\eta^T\psi(s)$, where $\psi(s) = [\psi_i(z)]_{i=1}^d\in \RR^{d}$, $\theta, \eta\in \RR^d$, our saddle point problem \eq{eq:RL_dual} reduces to 
$
\min_{\theta} \big\|\EE_{s, a, s'}[\Delta_\theta(s, a, s')\psi]\big\|^2_{\EE[\psi\psi^\top]^{-1}}, 
$
where $\Delta_\theta(s, a, s') = R(s, a) + \gamma V^\pi(s') - V^\pi(s)$. This is exactly the same as the objective proposed in~\cite{SutMaePreBhaetal09} of gradient-TD2. Moreover, the update rules in gradient-TD2 can also be derived by conducting the proposed Embedding-SGD with such parametrization. For details of the derivation, please refer to Appendix~\ref{appendix:GTD_special}.

From this perspective, gradient-TD2 is simply a special case of the proposed Embedding-SGD applied to policy evaluation with particular parametrization. However, in the view of our framework, there is really no need to, and should not, restrict to the same finite parametric model for the value and dual functions. As further demonstrated in our experiments, with different nonparametric models, the performances can be improved significantly. See details in Section~\ref{subsec:RL_exp}. 

The residual gradient~(RG)~\cite{Baird95} is trying to apply stochastic gradient descent directly to the MSBE with finite parametric form of value function, \ie, $V^\pi(s)=\theta^T\psi(s)$, resulting the gradient as 
$
\EE_{s, a, s'}\sbr{\Delta_\theta(s, a, s')\psi(s)} - \gamma\EE_s\sbr{\EE_{s', a|s}\sbr{\Delta_\theta(s, a, s')}\EE_{s'|s}\sbr{\psi(s')}}.
$
Due to the inside conditional expectation in gradient expression, to obtain an unbiased estimator of the gradient, it requires two independent samples of $s'$ given $s$, which is not practical. To avoid such ``double sample'' problem, \citet{Baird95} suggests to use gradient as $\EE_{s, a, s'}\sbr{\Delta_\theta(s, a, s')\rbr{\psi(s) - \gamma\psi(s')}}$. In fact, such algorithm is actually optimizing $\EE_{s, a, s'}\sbr{\Delta_\theta(s, a, s')^2}$, which is an upper bound of MSBE~\eq{eq:RL_obj} because of the convexity of square loss. 

Our algorithm is also fundamentally different from the TD algorithm even in the finite state case. The TD algorithm updates the state-value function directly by an estimate of the temporal difference based on one pair of samples, while our algorithm  updates the state-value function based on accumulated estimate of the temporal difference, which intuitively is more robust. 

\noindent{\bf Optimal control.} The goal is to estiamte the $z(\cdot)$ function by minimizing the error in linear Bellamn equation~\eq{eq:OC_obj}. With $z\in \tilde\Hcal$ with feature map $\psi(\cdot)$, we apply the dual kernel embedding, which will lead to the saddle point problem
\begin{eqnarray}\label{eq:OC_dual}
\min_{z \in\tilde{\Hcal}} \max_{u\in\Hcal} \Phi(z,u):=\EE_{s', s} \sbr{\langle z, \psi(s) -  \exp(-R(s))\psi(s') \rangle_{\tilde \Hcal}\cdot u(s)} -\frac{1}{2}\EE_{s}[u^2(s)].
\end{eqnarray}
With the learned $z^*$, we can recover the optimal control via its conditional distribution $p^{\pi_*}(s'|s) = \frac{p(s'|s)z^*(s)}{\EE_{s'|s\sim p(s'|s)}\sbr{z^*(s')}}$.

\subsection{Events Prediction in Stochastic Processes} 
{\bf Expected Hitting Time.} The goal is to estimate the expected hitting time $h(\cdot, \cdot)$ which minimizes the recursive error~\eq{eq:HT_obj}. With $h\in \tilde \Hcal$ with feature map $\psi(\cdot, \cdot)$, we apply the dual kernel embedding similarly to the policy evaluation, which will lead to the saddle point problem
\begin{eqnarray}\label{eq:HT_dual}
\min_{h \in\tilde{\Hcal}} \max_{u\in\Hcal} \Phi(h,u):=\EE_{s, t}\EE_{s'|s, t} \sbr{ \rbr{1 - \langle h,  \psi(s, t) - \psi(s', t) \rangle_{\tilde\Hcal} } u(s, t)}- \frac{1}{2}\EE_{s}\sbr{u(s, t)^2}.
\end{eqnarray}

\noindent{\bf Remark.} With the proposed algorithm, we can estimate $h(s, t)$, the expected hitting time starting from state $s$ and hitting state $t$, even without observing the hitting events actually happen. We only need to collect the trajectory of the stochastic processes, or just the one-step transition, to feed to the algorithm, therefore, utilize the data more efficietly.

\section{Experiments}

We test the proposed algorithm for two applications, \ie, learning with invariant representation and policy evaluation. For full details of our experimental setups, please refer to Appendix~\ref{appendix:experiment_setup}. 

\subsection{Experiments on Invariance Learning} 

To justify the algorithm for learning with invariance, we test the algorithm on two tasks. We first apply the algorithm to robust learning problem where the inputs are contaminated, and then, we conduct comparison on molecular eneretics prediction problem~\cite{MonHanFazRupetal12}. We compare the proposed algorithm with SGD with virtual samples technique~\cite{NiyGirPog98,LooCanBot07} and SGD with finite sample average for inner expectation~(SGD-SAA). We use Gaussian kernel in all tasks. To demonstrate the sample-efficiency of our algorithm, $10$ virtual samples are generated for each datum in training phase. The algorithms are terminated after going through $10$ rounds of training data. We emphasize that the SGD with virtual samples is optimizing an upper bound of the objective, dan thus, it is predictable that our algorithm can achieve better performance. We plot its result with dot line instead.

\begin{figure*}[!t]
\centering
  \begin{tabular}{cc}
    \includegraphics[width=0.35\textwidth]{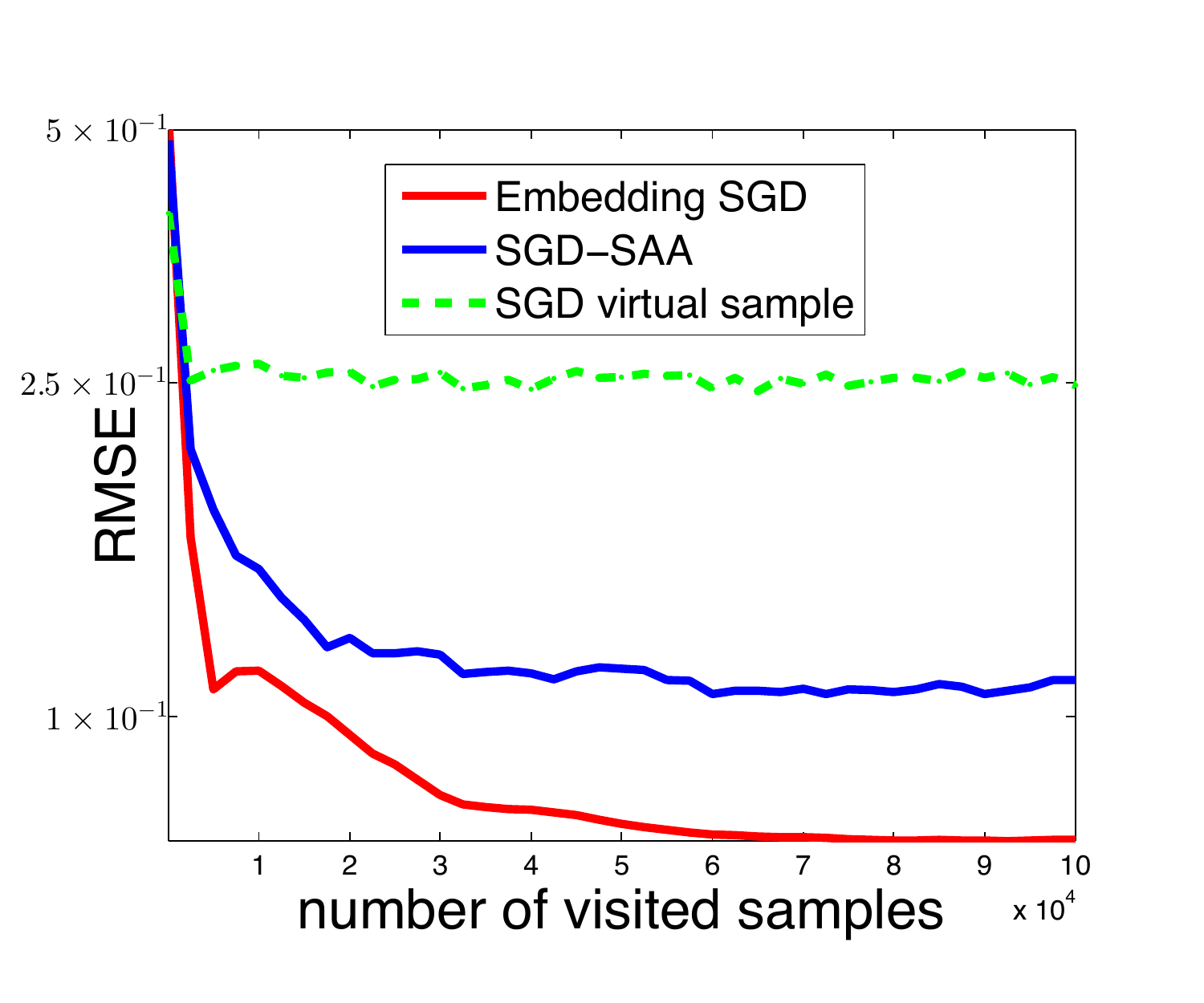}&~~~~~~~~~~~~
    \includegraphics[width=0.35\textwidth]{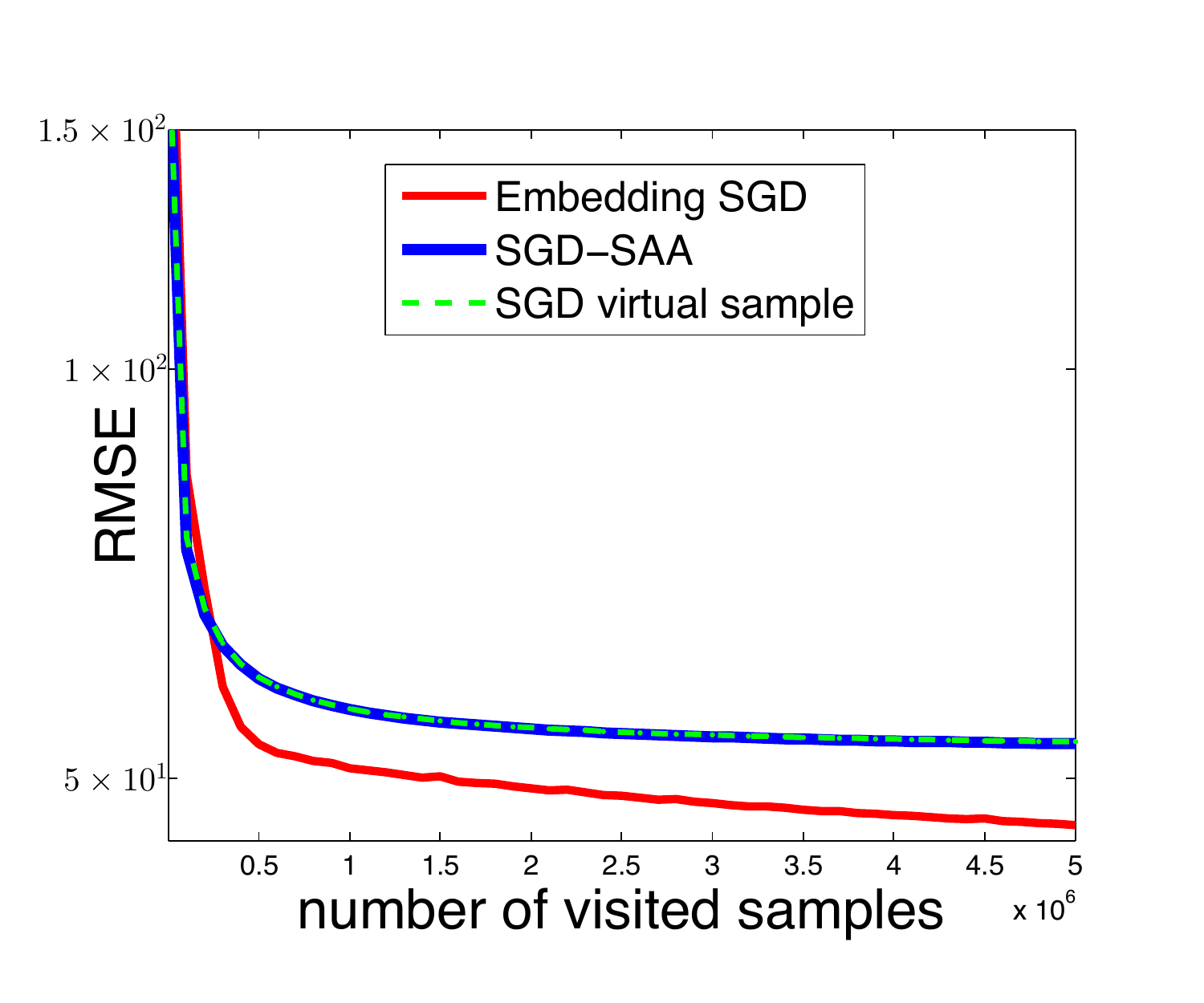}\\
    (a) Robust Learning &~~~~~~~~~~~~(b) QuantumMachine\\
  \end{tabular}
  \caption{Learning with invariance.}
  \label{fig:learning_invariant}
\end{figure*}

\paragraph{\bf Noisy measurement.} We generate a synthetic dataset by
\begin{eqnarray*}
\bar x &\sim&\Ucal([-0.5, 0.5]),\quad  x = \bar x + 0.05e,\\
y&=& (\sin(3.53\pi \bar x) + \cos(7.7 \pi \bar x))\exp(-1.6\pi|\bar x|) + 3 \bar x^2+ 0.01e,
\end{eqnarray*}
where the contamination $e\sim \Ncal(0, 1)$. Only $(x, y)$ are provided to learning methods, while $\bar x$ is unknown. The virtual samples are sampled from $z\sim \Ncal(x, 0.05^2)$ for each observation. The 10 runs average results are illustrated in Figure~\ref{fig:learning_invariant}(a). The proposed algorithm achieves average MSE as low as $0.0029$ after visit $0.1$M data, significantly better than the alternatives.

\paragraph{\bf QuantumMachine.} We test the proposed algorithm for learning with invariance task on QuantumMachine 5-fold dataset for atomization energy prediction. We follow~\cite{MonHanFazRupetal12} that the data points are represented by Coulomb matrices, and the virtual samples are generated by random permutation. The average results are shown in Figure~\ref{fig:learning_invariant}(b). The proposed algorithm achieves a significant better solution, while SGD-SAA and SGD with virtual samples stuck in inferior solutions due to the inaccurate inner expectation estimation and optimizing indirect objective, respectively.

\subsection{Experiments on Policy Evaluation}\label{subsec:RL_exp}

We compare the proposed algorithm to several prevailing algorithms for policy evaluation, including gradient-TD2~(GTD2)~\cite{SutMaePreBhaetal09,LiuLiuGhaMah15}, residual gradient~(RG)~\cite{Baird95} and kernel MDP~\cite{GruLevBalPonetal12} in terms of mean square Bellman error~\cite{DanGerPet14}. It should point out that kernel MDP is not an online algorithm, since it requires to visit the entire dataset when estimating the embedding and inner expectation in each iteration. We conduct experiments for policy evaluation on several benchmark datasets, including navigation, cart-pole swing up and PUMA-560 manipulation. We use Gaussian kernel in the nonparametric algorithms, \ie, kernel MDP and Embedding SGD, while we test random Fourier features~\cite{RahRec08} for the parametric competitors, \ie, GTD2 and RG. In order to demonstrate the sample efficiency of our method, we only use one sample from the conditional distribution in the training phase, therefore, cross-validation based on Bellman error is not appropriate. We perform a parameter sweep to select the hyper-parameters as~\cite{SilLevHeeetal14}. See appendix \ref{appendix:experiment_setup} for detailed settings. Results are averaged over 10 independent trials.
\begin{figure*}[!t]
\centering
  \begin{tabular}{ccc}
    \includegraphics[width=0.312\textwidth]{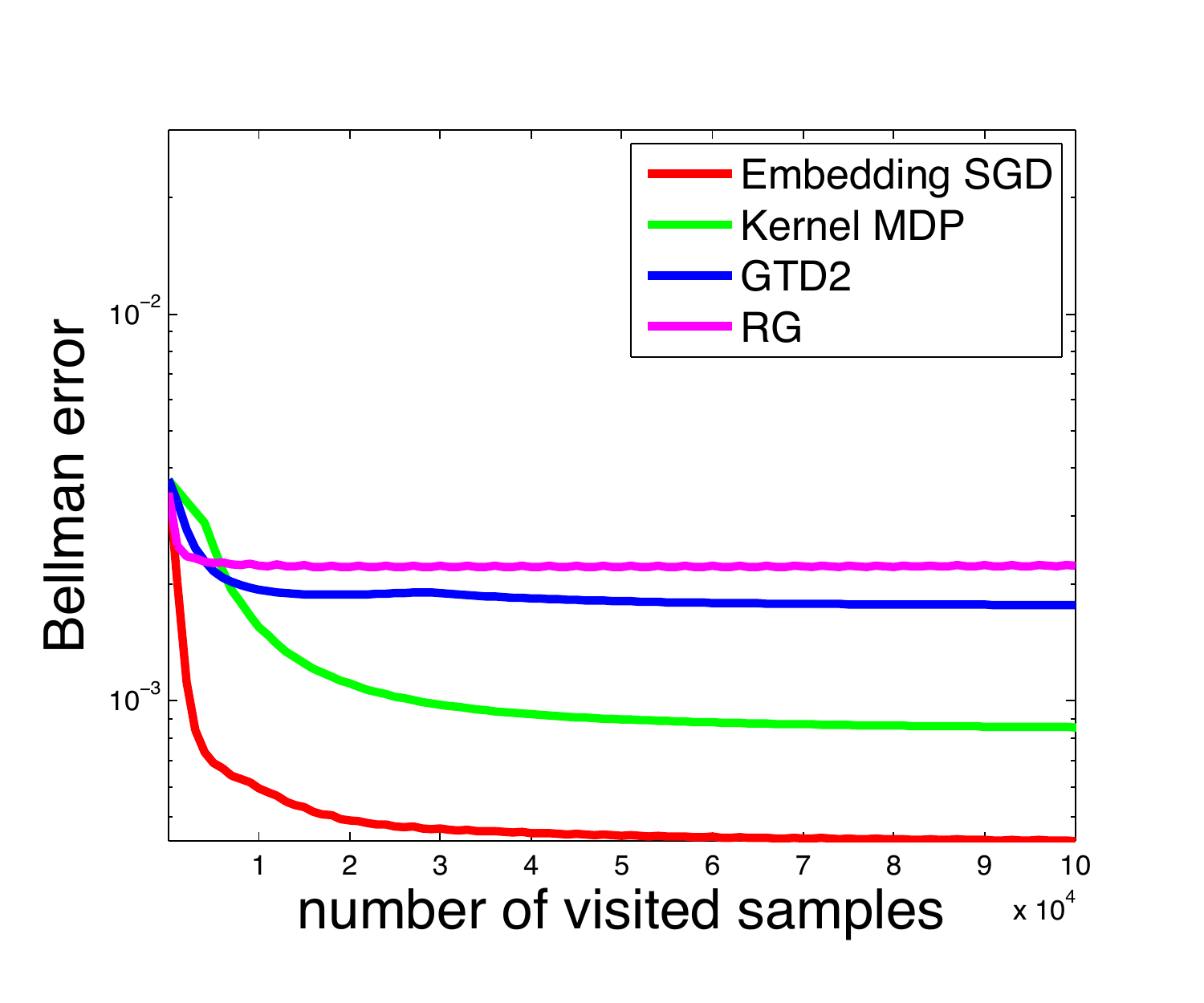}&
    \includegraphics[width=0.312\textwidth]{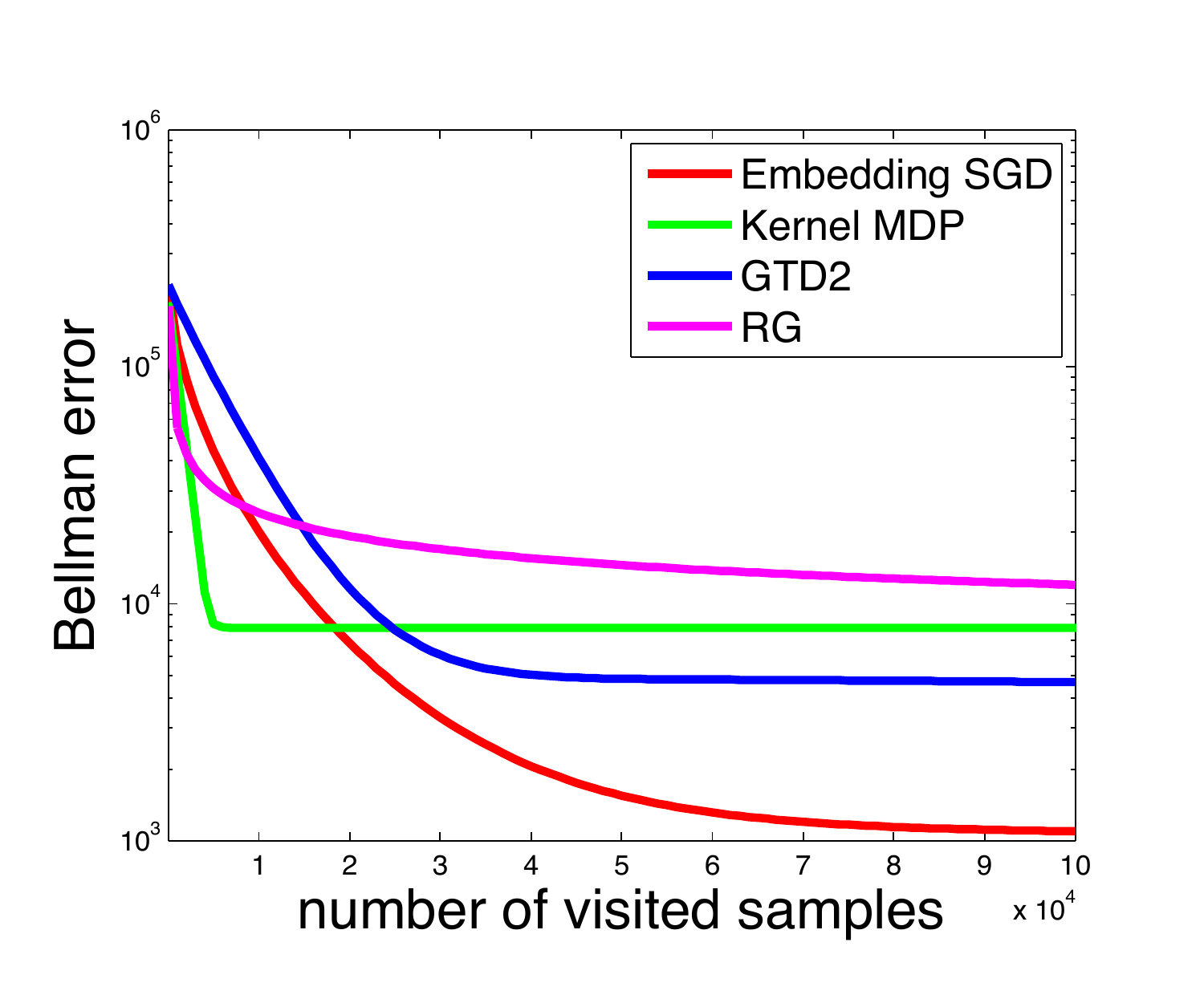}&
    \includegraphics[width=0.312\textwidth]{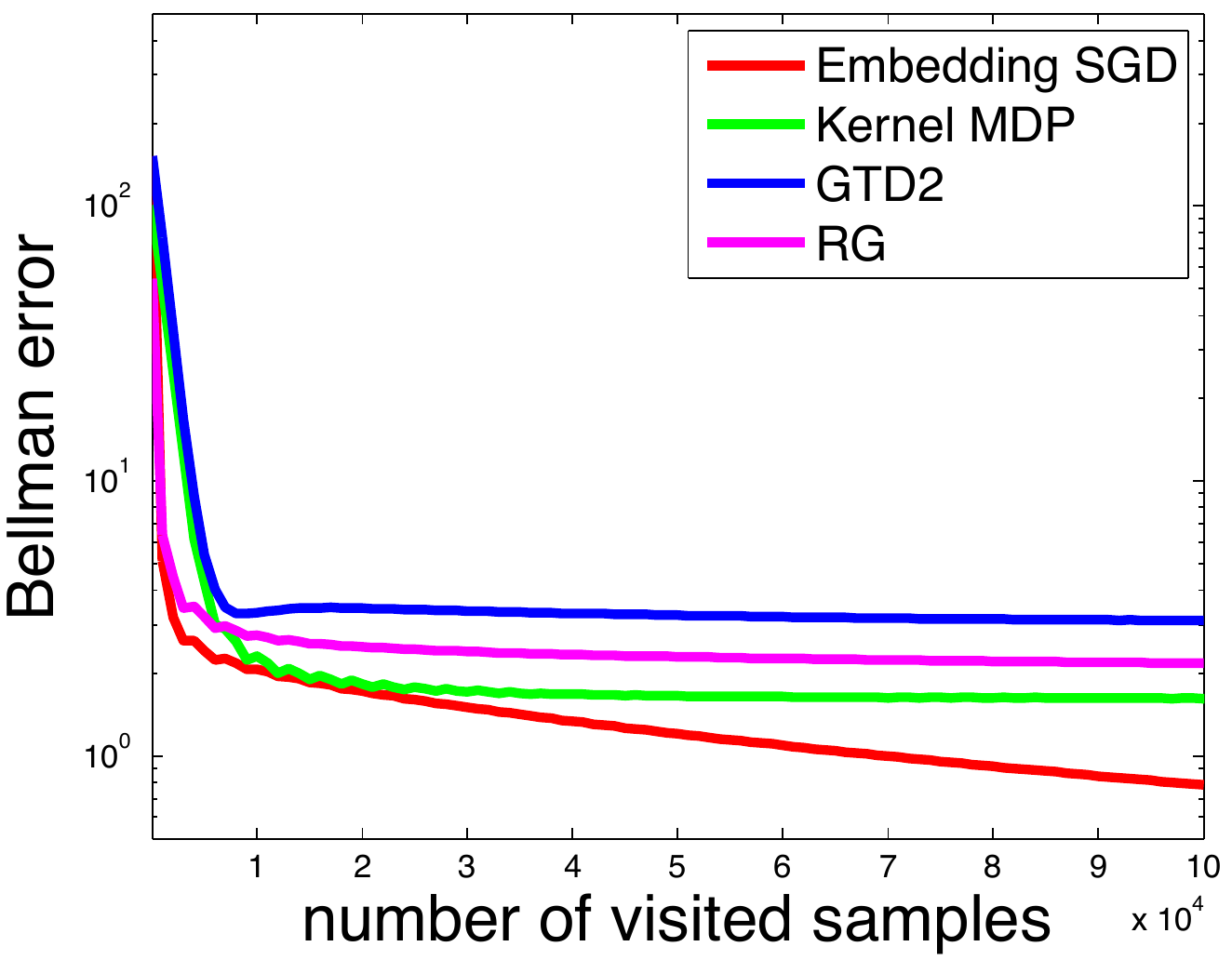} \\
    (a) Navigation & (b) Cart-Pole &(c) PUMA-560\\
  \end{tabular}
  \caption{Policy evaluation.}
  \label{fig:policy_evaluation}
\end{figure*}

\paragraph{\bf Navigation.} The navigation in an unbounded room task extending the discretized MDP in~\cite{GruLevBalPonetal12} to continuous-state continuous-action MDP. Specifically, the reward is $R(s) = \exp(-100\|s\|^2)$ centered in the middle of the room and $s\sim \Ncal(0, 0.2I)$. We evaluate the deterministic policy policy $\pi(s) = -0.2s R(s)$, following the gradient of the reward function. The transition distribution follows Gaussian distribution, $ p(s'|a, s) = \Ncal(s+a, 0.1I)$. Results are reported in Figure~\ref{fig:policy_evaluation}(a).

\paragraph{\bf Cart-pole swing up.} The cart-pole system consists of a cart and a pendulum. It is an under-actuated system with only one control act on the cart. The goal is to swing-up the pendulum from the initial position (point down). The reward will be maximum if the pendulum is swing up to $\pi$ angle with zero velocity. We evaluate the linear policy $\pi(s) = As+b$ where $A\in \mathbb{R}^{1\times 4}$ and $b\in\mathbb{R}^{1\times 1}$. Results are reported in Figure~\ref{fig:policy_evaluation}(b). 

\paragraph{\bf PUMA-560 manipulation.} PUMA-560 is a robotic arm that has 6 degrees of freedom with  6 actuators on each joint. The task is to steer the end-effector to the desired position and orientation with zero velocity. The reward function is maximum if the arm is located to the desired position. We evaluate the linear policy $\pi(s) = {A}s+{b}$ where ${A}\in \mathbb{R}^{6\times 12}$ and ${b}\in\mathbb{R}^{6\times 1}$. Results are reported in Figure~\ref{fig:policy_evaluation}(c). 

In all experiments, the proposed algorithm performs consistently better than the competitors. The advantages of proposed algorithm mainly come from three aspects: {\bf i)}, it utilizes more flexible dual function space, rather than the constrained space in GTD2; {\bf ii)}, it directly optimizes the MSBE, rather than its surrogate as in GTD2 and RG; {\bf iii)}, it directly targets on value function estimation and forms an one-shot algorithm, rather than a two-stage procedure in kernel MDP including estimating conditional kernel embedding as an intermediate step.

\section{Conclusion}\label{sec:conclusion}
We propose a novel \emph{sample-efficient} algorithm, {\bf Embedding-SGD}, for addressing learning from conditional distribution problems. Our algorithm benefits from a fresh employ of saddle point and kernel embedding techniques, to mitigate the difficulty with limited samples from conditional distribution as well as the presence of nested expectations. To our best knowledge,  among all existing algorithms able to solve such problems, this is \emph{the first} algorithm that allows to take only one sample at a time from the conditional distribution and comes with provable theoretical guarantee. 

We apply the proposed algorithm to solve two fundamental problems in machine learning, \ie, learning with invariance and policy evaluation in reinforcement learning. The proposed algorithm achieves the state-of-the-art performances on these two tasks comparing to the existing algorithms. As we discussed in Appendix~\ref{appendix:control}, the algorithm is also applicable to control problem in reinforcement learning. 

In addition to its wide applicability,  our algorithm is also very versatile and amenable for all kinds of enhancement. The algorithm can be easily extended with random feature approximation or doubly stochastic gradient trick. Moreover, we can extend the framework by learning complicated nonlinear feature jointly with the function, which results the dual neural network embedding in Appendix~\ref{appendix:dual_neural_networks}. It should be emphasized that since the primal and dual function spaces are designed for different purposes, although we use both RKHS in main text for simplicity, we can also use different function approximators separately for primal and dual functions.


\bibliographystyle{plainnat}

{
}

\clearpage
\newpage

\appendix
\onecolumn

\begin{appendix}

\thispagestyle{plain}
\begin{center}
{\Large \bf Appendix}
\end{center}

\section{Interchangeability Principle and Dual Continuity}\label{appendix:dualcontinuity}

\textbf{Lemma~\ref{lem:switch_correct}} {\it
Let $\xi$ be a random variable on $\Xi$ and assume  for any $\xi\in \Xi$, function $g(\cdot,\xi):\RR\to(-\infty,+\infty)$ is a proper and upper semicontinuous concave function. Then
\begin{equation*}
\EE_{\xi}[\max_{u\in\RR}g(u,\xi)] =\max_{u(\cdot)\in \Gcal(\Xi)}\EE_{\xi}[g(u(\xi),\xi)]. 
\end{equation*}
where $\Gcal(\Xi)=\{u(\cdot):\Xi\to\RR\}$ is the entire space of functions defined on support $\Xi$. \\  
}

\begin{proof}
First of all, by assumption of concavity and upper-semicontinuity, we know that for any $\xi\in\Xi$, there exists a maximizer for $\max_u g(u,\xi)$; let us denote as $u^*_{\xi}$. We can therefore define a function $u^*(\cdot):\Xcal\to\RR$ such that $u^*(\xi)=u^*_\xi$, and thus, $u^*(\cdot)\in \Gcal(\Xi)$. Hence, 
\begin{equation*}
\EE_{\xi}[\max_{u\in\RR}g(u,\xi)] =\EE_{\xi}[g(u^*(\xi),\xi)]\leq\max_{u(\cdot)\in \Gcal(\Xcal)}\EE_{\xi}[g(u(\xi),\xi)]. 
\end{equation*}
On the other hand, clearly, for any $u(\cdot)\in\Gcal(\Xi)$ and $\xi\in\Xi$, $g_\xi(u(\xi),\xi)\leq \max_{u\in\RR} g(u,\xi)$. Hence, $\EE_{\xi}[g(u(\xi),\xi)]\leq \EE_{\xi}[\max_{u\in\RR} g(u,\xi)]$, for any $u(\cdot)\in \Gcal(\Xi)$. This further implies that  
\begin{equation*}
\max_{u(\cdot)\in \Gcal(\Xi)}\EE_{\xi}[g(u(\xi),\xi)]\leq \EE_{\xi}[\max_{u\in\RR}g(u,\xi)]. 
\end{equation*}
Combining these two facts leads to the statement in the lemma. 
\end{proof}
\\
\textbf{Proposition~\ref{prop:dualcontinuity}} {\it
Suppose both  $f(z,x)$ and $p(z|x)$ are continuous in $x$ for any $z$,
\begin{enumerate}
\item[(1)] (Discrete case) If the loss function $\ell_y(v)$ is continuously differentiable in $v$ for any $y\in\Ycal$, then $u^*(x,y)$ is unique and continuous in $x$  for any $y\in \Ycal$;
\item[(2)] (Continuous case) If the loss function $\ell_y(v)$ is continuously differentiable in $(v,y)$, then $u^*(x,y)$ is unique and continuous in $(x,y)$  on $\Xcal\times\Ycal$. 
\end{enumerate}
}
\begin{proof}
The continuity properties of optimal dual function follows directly from the fact that $u^*(x,y)\in \partial \ell_y(\EE_{z|x}[f(z,x)])$. In both cases, for any $y\in \Ycal$, $\ell_y(\cdot)$ is differentiable. Hence $u^*(x,y)=\ell_y'(\int f(z,x)p(z|x)dz)$ is unique.  Since $f(z,x)$ and $p(z|x)$ is continuous in $x$ for any $z$, then $\EE_{z|x}[f(z,x)]$ is continuous in $x$. Since for any $y\in\Ycal$, $\ell'_y(\cdot)$ is continuous, the composition $u^*(x,y)$ is therefore continuous in $x$ as well. Moreover, if $\ell'_y(\cdot)$ is also continuous in $y\in\Ycal$, then the composition $u^*(x,y)$ is continuous in $(x,y)$.  
\end{proof}
Indeed, suppose $\ell_y(\cdot)$ is uniformly $L$-Lipschitz differentiable for any $y\in \Ycal$, $f(z,x)$ is uniformly $M_f$-Lipschitz continuous in $x$ for any $z$, $p(z|x)$ is $M_p$-Lipschitz continuous in $x$. Then 
\begin{align*}
 |u^*(x_1,y)-u^*(x_2,y)|&=\bigg|\ell_y'\rbr{\int f(z,x_1)p(z|x_1)dz}-\ell_y'\rbr{\int f(z,x_2)p(z|x_2)dz}\bigg|\\
 &\leq L\int|f(z,x_1)p(z|x_1)-f(z,x_2)p(z|x_2)|dz\\
  &\leq L\int|f(z,x_1)-f(z,x_2)|p(z|x_1)dz +L\int|f(z,x_2)|\cdot|p(z|x_1)-p(z|x_2)|dz\\
&\leq LM_f|x_1-x_2|+LM_p|x_1-x_2|\sup_{x}\int |f(z, x)|dz
\end{align*}
If for any $f(z,x)$ is Lebesgue integrable and $\int|f(z,x)|dz$ is uniformly bounded, then $u^*(x,y)$ is also Lipschitz-continuous for any $y\in \Ycal$. Moreover, if in addition, $\ell_y(v)$ is also Lipschitz differentiable in $(v,y)$, then $u^*(x,y)$ is also Lipschitz continuous on $\Xcal\times \Ycal$.

\section{Preliminaries: Stochastic Approximation for Saddle Point Problems}\label{appendix:prellim_convergece_rate}
Consider the stochastic saddle point (min-max) problem
$$\min_{x\in X}\max_{y\in Y}\Phi(x,y)=\EE[F(x,y,\xi)] $$
where the expected value function $f(x,y)$ is convex in $x$ and concave in $y$, and domains $X,Y$ are convex closed. Let $z=[x,y]$ and $G(z,\xi)=[\nabla F_x(x,y,\xi);-\nabla F_y(x,y,\xi)]$ be the stochastic gradient for any input point $z$ and sample $\xi$. Let $\|\cdot\|$ be a norm defined on the embedding Hilbert space of $Z=X\times Y$, and  $D(z,z'):=w(z)-w(z')-\nabla w(z')'(z-z')$ be a Bregman distance on $Z$ defined by a $1$-strongly convex (w.r.t. the norm $\|\cdot\|$) and continuously differentiable function $w(z)$. For instance, when $w(z)=\frac{1}{2}\|z\|^2$, the Bregman distance becomes $D(z,z')=\frac{1}{2}\|z-z'\|^2$.

\paragraph{Mirror descent SA.} The mirror descent stochastic approximation~\cite{NemJudLanSha09} works as follows:
$$ z_i = \argmin_{z\in Z} \{D(z, z_i)+\gamma_i G(z_i,\xi_i)\}, i = 1, \ldots, t.$$
The quality of an approximate solution $\bar z=(\bar x,\bar y)$ is defined by the error

$$\epsilon_{\rm gap}(\bar x,\bar y):= \max_{y\in Y}\Phi(\bar x,y)- \Phi^* + \Phi^* - \min_{x\in X}\Phi(x,\bar y) =\max_{y\in Y}\Phi(\bar x,y)-\min_{x\in X}\Phi(x,\bar y),$$
where $\Phi^*$ denotes the optimal value. Let $\bar z_t:=\frac{\sum_{i=1}^t\gamma_i z_i}{\sum_{i=1}^t\gamma_i}$, the convergence properties of this weighted averaging solution is as follows.
\begin{lemma}\label{lem:SA}
\cite{NemJudLanSha09} Suppose $\EE[\|G(z_i,\xi_i)\|_*^2]\leq M^2, \forall i$, we have
$$ \EE[\epsilon_{\rm gap}(\bar x_t, \bar y_t)]\leq \frac{2\max_{z\in Z}D(z, z_1)+\frac{5}{2}M^2\sum_{i=1}^t\gamma_i^2}{\sum_{i=1}^t\gamma_i}.$$
\end{lemma}

In particular, when $\gamma_i=\frac{\gamma}{\sqrt{t}}, \forall i=1,\ldots, t$, we have
$$\EE[\epsilon_{\rm gap}(\bar x_t, \bar y_t)]\leq (2\max_{z\in Z}D(z, z_1)/\gamma +\frac{5}{2}M^2\gamma)\frac{1}{\sqrt{t}}.$$
Moreover, suppose $D^2=\max_{z\in Z}D(z, z_1)$ and $M^2$ are known, by setting $\gamma=\frac{2D}{\sqrt{5}M}$, we further have
$$\EE[\epsilon_{\rm gap}(\bar x_t, \bar y_t)]\leq \frac{2\sqrt{5}DM}{\sqrt{t}}.$$
To summarize, the mirror descent stochastic approximation achieves an $\Ocal(1/\sqrt{t})$ convergence rate (also known to be unimprovable \cite{NemJudLanSha09}). Our Embedding-SGD algorithm~\ref{alg:stochastic_composite_general} builds upon on this framework to solve the saddle point approximation problem (\ref{eq:dual_approximate}).

\section{Convergence Analysis for Embedding-SGD}\label{appendix:convergece_rate}

\subsection{Decomposition of generalization error} 
Let ${f}_*$ be the optimal solution to our objective. Denote $\hat L(f)=\max_{u\in\Hcal^\delta}\phi(f,u)$. Invoking the Lipschitz continuity of $\Phi$ ,
$L(f)-\hat L(f)\leq (K+C)\Ecal(\delta), \forall f$. 
Therefore, 
\begin{align*}
L(\bar {f}_t)-L({f}_*)=&L(\bar{f}_t)-\hat L(\bar{f}_t)+\hat L(\bar{f}_t)-\hat L({f}_*)+\hat L({f}_*)-L({f}_*)\\
&\leq\epsilon_{\rm gap}(\bar{f}_t,\bar{u}_t) +2(K+C)\Ecal(\delta).
\end{align*}

\subsection{Optimization error}
\paragraph{Proof of Theorem~\ref{thm:main}}
Our proof builds on results of stochastic approximation discussed in the previous section. Let $M_1$ and $M_2$ be such that for any ${f}\in\{{f}_i\}_{i=1}^t$ and $u\in\{u_i\}_{i=1}^t$,
$$\EE_{x,y,z}[\|\nabla_{{f}}\hat\Phi_{x,y,z}({f},u)\|_\Fcal^2]\leq M_1^2$$
$$\EE_{x,y,z}[\|\nabla_{u}\hat\Phi_{x,y,z}({f},u)\|_\Hcal^2]\leq M_2^2$$
Then from Lemma~\ref{lem:SA}, we have
\begin{equation}\label{eq:bound}
\EE[\epsilon_{\rm gap} (\bar{f}_t,\bar  y_t)]\leq \frac{2(D_{\Fcal}^2+D_\Hcal^2)+\frac{5}{2}(M_1^2+M_2^2)\sum_{i=1}^t\gamma_i^2}{\sum_{i=1}^t\gamma_i}
\end{equation}
where $D_{\Fcal}^2=\sup_{{f}\in{\Fcal}}\frac{1}{2}\|{f}_1-{f}\|_2^2$ and $D_\Hcal^2=\sup_{u\in\Hcal^\delta}\frac{1}{2}\|u_1-u\|_\Hcal^2\leq 2\delta$. It remains to find upper bounds for $M_1$ and $M_2$. Note that since $\|k(w,w')\|_\infty\leq \kappa$ for any $w$ and $w'$,
\begin{equation*}
\begin{array}{rl}
\EE[\|\nabla_{u}\hat\Phi_{x,y,z}({f},u)\|_\Hcal^2]&\leq\kappa\EE[\|f(z,x)-\nabla \ell_y^*(u(x))\|^2]\leq 2\kappa (M_\Fcal+c_\ell).
\end{array}
\end{equation*}
Since $u(x)=\langle u(\cdot),k(x,\cdot)\rangle_\Hcal$, from Young's inequality, we have
$|u(x)|\leq \frac{1}{2}\|u\|_\Hcal^2+\frac{1}{2}\|k(x,\cdot)\|_\Hcal^2\leq \frac{1}{2}(\delta+\kappa)$, for any $w\in\Xcal$.
\begin{equation*}
\begin{array}{rl}
\EE_{x,y,z}[\|\nabla_{{f}}\hat\Phi_{x,y,z}({f},u)\|_\Fcal^2]&=\EE[\|\psi(z,x)\|_\Fcal^2 u(x)^2]\leq \frac{1}{4}(\delta+\kappa)^2C_\Fcal.
\end{array}
\end{equation*}
Plugging in $M_1^2=2\kappa (M_\Fcal+c_\ell)$ and $M_2^2=\frac{1}{4}(\delta+\kappa)^2C_\Fcal$ to (\ref{eq:bound}) and setting $\gamma_t=\gamma/\sqrt{t}$, we arrive at (\ref{eq:mainbound}).
\QED

\section{Gradient-TD2 As Special Case of Embedding-SGD}\label{appendix:GTD_special}

Follow the notation in section~\ref{sec:application}, with the parameterization that $V^\pi(s)=\theta^T\psi(s)$ and $u(s)=\eta^T\psi(s)$, where $\psi(s) = [\psi_i(z)]_{i=1}^d\in \RR^{d}$, $\theta, \eta\in \RR^d$, the optimization becomes 
\begin{eqnarray}\label{eq:GTD2_opt}
\min_{\theta} \max_{\eta } \widehat\Phi(\theta, \eta):=\EE_{s}\EE_{s',a|s} \Big[\Delta_\theta(s, a, s') \psi(s)^\top\eta\Big] -\frac{1}{2}\EE_{s}[\eta^\top\psi(s)\psi^\top(s)\eta],
\end{eqnarray}
where $\Delta_\theta(s, a, s') = \rbr{R(s) + \gamma \theta^\top \psi(s') - \theta^\top \psi(s)}$. For arbitrary $\theta$, we have the closed form of $\eta(\theta)^*$ which achieves the maximum of $\widehat \Phi(\theta, \eta)$. Specifically, we first take derivative of $\widehat\Phi(\theta, \eta)$ w.r.t. $\eta$, 
\begin{eqnarray}
\nabla_\eta \widehat\Phi(\theta, \eta) = \EE_{s, a, s'}\sbr{\Delta_\theta(s, a, s')\psi(s)} - \EE_{s}\sbr{\psi(s)\psi(s)^\top \eta},
\end{eqnarray}
and make the derivative equal to zero,
\begin{eqnarray}
\eta(\theta)^* = \EE_s\sbr{\psi(s)\psi(s)^\top }^{-1}\EE_{s, a, s'}\sbr{\Delta_\theta(s, a, s')\psi(s)}.
\end{eqnarray}
Plug the $\eta(\theta)^*$ into $\widehat \Phi(\theta, \eta)$, we achieve the optimization
\begin{eqnarray}
\min_\theta \EE_{s, a, s'}\sbr{\Delta_\theta(s, a, s')\psi(s)^\top}\EE_s\sbr{\psi(s)\psi(s)^\top }^{-1}\EE_{s, a, s'}\sbr{\Delta_\theta(s, a, s')\psi(s)},
\end{eqnarray}
which is exactly the objective of gradient-TD2~\cite{SutMaePreBhaetal09,LiuLiuGhaMah15}. Plug the parametrization into the proposed embedding-SGD, we will achieve the update rules in $i$-th iteration proposed in gradient-TD2 for $\theta$ and $\eta$ as
\begin{equation*}
\begin{array}{rcl}
\eta_{i+1} &=&\eta_i+\gamma_i[\Delta_\theta(s, a, s')-u_i(s)]\psi(s),\\
\theta_{i+1} &= & \theta_i-\gamma_iu_i(s)(\gamma \psi({s'})- \psi(s)).\\
\end{array}
\end{equation*}
Therefore, from this perspective, gradient-TD2 is simply a special case of the proposed Embedding-SGD applied to policy evalution with particular parametrization.

\section{Dual Embedding with Arbitrary Funtion Approximator}\label{appendix:extend_dual_embedding}

In the main text, we only focus on using different RKHSs as the primal and dual function spaces. As we introduce in section~\ref{sec:intro}, the proposed algorithm is versatile and can be conducted with arbitrary function space for the primal or dual functions. In this section, we demonstrate applying the algorithm to random feature represented functions~\cite{RahRec08} and neural networks. For simplicity, we specify the algorithms with either kernel, random feature representation or neural networks for both primal and dual functions. It should be emphasized that in fact the parametrization choice of the dual function is \emph{independent} to the form of the primal function. Therefore, the algorithm can also be conducted in \emph{hybrid setting} where the primal function uses one form of function approximator, while the dual function use another form of function approximator.

Instead of solving (\ref{eq:dual_approximate}), in this section, we consider the alternative reformulation by penalizing the norm of the dual function, which has been widely used as an alternative to the constrained problem in machine learning literatures, and is proven to be more robust often times in practice,
\begin{equation}\label{eq:dual_regularized}
 \min_{f\in\Fcal} \max_{u\in \Hcal} \;\Phi(f,u) + \frac{\lambda_1}{2}\|f\|_\Fcal^2-\frac{\lambda_2}{2}\|u\|_\Hcal^2
\end{equation}
It is well-known that there is a one-to-one relation between $\delta_\Fcal$, $\delta$ and $\lambda_1$, $\lambda_2$, respectively, such that the optimal solutions to (\ref{eq:dual_approximate}) and (\ref{eq:dual_regularized}) are the same. The objective can also be regarded as a smoothed approximation to the original problem of our interest, see \cite{Nesterov05}. Problem (\ref{eq:dual_regularized}) can be solved efficiently via our Algorithm~\ref{alg:stochastic_composite_general} by simply revoking the projection operators. 

\subsection{Dual Random Feature Embeddings}\label{appendix:dual_random_fea}

In this section, we specify the proposed algorithm leveraging random feature to approximate kernel function. For arbitrary positive definite kernel, $k(x, x)$, there exists a measure $\PP$ on $\Xcal$, such that $k(x, x') = \int \widehat\phi_w(x)\widehat\phi_w(x')d\PP(w)$~\cite{Devinatz53,HeiBou04}, where random feature $\widehat\phi_w(x):\Xcal\rightarrow \RR$ from $L_2(\Xcal, \PP)$. Therefore, we can approximate the function $f\in \Hcal$ with Monte-Carlo approximation $\hat f \in \widehat\Hcal^m=\{\sum_{i=1}^m\beta_i\widehat\phi_{\omega_i}(\cdot)|\|\beta\|_2\leq C\}$ where $\{w_i\}_{i=1}^m$ sampled from $\PP(\omega)$~\cite{RahRec09}. With such approximation, we obtain the corresponding \emph{dual random feature embeddings} variants. 

Denote the random feature for $\tilde k(\cdot, \cdot)$ and $k(\cdot, \cdot)$ as $\widehat\psi_w(\cdot)$ and $\widehat\phi_w(\cdot)$ with respect to distribution $\widetilde\PP(\omega)$ and $\PP(\omega)$, respectively, we approximate the $f(\cdot)$ and $u(\cdot)$ by $\hat f(\cdot)= \theta^\top \widehat\psi(\cdot)$ and $\hat u(\cdot ) = \eta^\top \widehat\phi(\cdot)$, where $\theta\in \RR^{m\times 1}$, $\eta\in \RR^{p\times 1}$, $\widehat\psi(\cdot) = [\widehat\psi_{\tilde w_1}(\cdot), \widehat\psi_{\tilde w_2}(\cdot), \ldots, \widehat\psi_{\tilde w_m}(\cdot)]^\top $ with $\{\tilde w_i\}_{i=1}^m\sim \widetilde\PP(\omega)$ and $\widehat\phi(\cdot) = [\widehat\phi_{w_1}(\cdot), \widehat\phi_{w_2}(\cdot), \ldots, \widehat\phi_{w_m}(\cdot)]^\top$ with $\{w_i\}_{i=1}^p\sim \PP(\omega)$. Then, we have the saddle point reformulation of~\eq{eq:target},
\begin{eqnarray}\label{eq:dual_random_fea}
\min_{\theta}\max_{\eta} \widehat\Phi(\theta, \eta):= \EE_{x, y}\EE_{z|x}\sbr{\theta^\top\widehat\psi(z, x) \widehat\phi(x,y)^\top \eta -l_y^*(\eta^\top\widehat\phi(x,y))} + \frac{\lambda_1}{2}\|\theta\|^2 - \frac{\lambda_2}{2}\|\eta\|^2.
\end{eqnarray}
Apply the proposed algorithm to~\eq{eq:dual_random_fea}, we obtain the update rule in $i$-th iteration,
\begin{equation*}
\begin{array}{rcl}
\theta_{i+1} &=& {(1 - \gamma_i\lambda_1)\theta_i-\gamma_i \hat u(x_i,y_i)\widehat \psi({z_i, x_i})},\\
\eta_{i+1} &=& (1 - \gamma_i\lambda_1)\eta_i + \gamma_i \big[\hat f_i(z_i, x_i) - \nabla \ell_{y_i}^*(\hat u(x_i,y_i))]\widehat \phi(x_i,y_i).\\
\end{array}
\end{equation*}

We emphasize that with the random feature representation will introduce an extra approximation error term in the order of  $\Ocal(1/\sqrt{m})$. To balance the approximate error and the statistical generalization error, we must use $m$ sufficiently large. 

\subsection{Extension to Embedding Doubly-SGD}\label{appendix:doublySGD}

To alleviate the approximation error introduced by random feature representation, we can further generalize the algorithmic technique about doubly stochastic gradient to the saddle point problem (\ref{eq:dual_regularized}), which can be viewed as setting $m$ to be infinite conceptually, therefore, eliminate the approximation error due to random feature representation. The embedding doubly-SGD is illustrated in Algorithm~\ref{alg:doublySGD},

\begin{minipage}{0.5\textwidth}
\begin{algorithm}[H]
\caption{\textbf{Embedding-Doubly SGD} for (\ref{eq:dual_regularized})}
  \text{\bf Input:} $\PP(x,y),\, \PP(z|x),\, \PP(\omega),\, \{\gamma_i\geq 0\}_{i=1}^t$\\[-4mm]
  \begin{algorithmic}[1]\label{alg:doublySGD}
    \FOR{$i=1,\ldots, t$}
      \STATE Sample $x_i,y_i \sim \PP(x,y)$.
      \STATE Sample $z_i \sim \PP(z|x_i)$.
      \STATE Sample $\omega_i\sim \PP(\omega)$ with seed $i$
      \STATE Sample $\widetilde\omega_i\sim \widetilde\PP(\omega)$ with seed $\tilde i$
      \STATE Compute $f_i=\textbf{Predict}(z_i, x_i, \{\alpha_j\}_{j=1}^{i})$
      \STATE Compute $u_i=\textbf{Predict}(x_i, y_i, \{\beta_j\}_{j=1}^{i})$
      \STATE $\alpha_{i+1} = \gamma_i u_i(x_i,y_i)\widehat\psi_{\widetilde\omega_i}(z_i, x_i)$.
      \STATE $\beta_{i+1} = \gamma_i [f_i(z_i,x_i)-\nabla\ell_{y_i}^*(u_i(x_i,y_i))]\widehat\phi_{\omega_i}(x_i,y_i) $.
      \STATE for $j=1,\ldots, i$\\
      $\alpha_j = (1-\gamma_i\lambda_1)\alpha_j$, $\beta_{j} = (1-\gamma_i\lambda_2)\beta_j$
    \ENDFOR\\
      \end{algorithmic}
\end{algorithm}
\end{minipage}
~~~
\begin{minipage}{0.4\textwidth}
  \begin{algorithm}[H]
  \caption{$u=\textbf{Predict}(x,y,\cbr{\beta_i}_{i=1}^t)$}
  \text{\bf Require: $\PP(\omega),\, \widehat\phi_{\omega}(x,y).$}\\[-4mm]
  \begin{algorithmic}[1] \label{alg:dual_predict}
    \STATE Set $u = 0$.
    \FOR{$i=1,\ldots, t$}
      \STATE Sample $\omega_i \sim \PP(\omega)$ with seed $i$.
      \STATE $u = u+ \beta_i \widehat\phi_{\omega_i}(x,y)$.
    \ENDFOR
  \end{algorithmic}
  \end{algorithm}
  \vspace{-5mm}
  \begin{algorithm}[H]
  \caption{$f=\textbf{Predict}(z, x,\,\cbr{\alpha_i}_{i=1}^t)$}
  \text{\bf Require: $\widetilde\PP(\omega),\, \widehat\psi_{\omega}(z, x).$}\\[-4mm]
  \begin{algorithmic}[1] \label{alg:primal_predict}
    \STATE Set $f = 0$.
    \FOR{$i=1,\ldots, t$}
      \STATE Sample $\widetilde\omega_i \sim \widetilde\PP(\omega)$ with seed $\tilde i$.
      \STATE $f = f+ \alpha_i \widehat\psi_{\widetilde\omega_i}(z,,x)$.
    \ENDFOR
  \end{algorithmic}
  \end{algorithm}
\end{minipage}

\vspace{3mm}

\subsection{Dual Neural Networks Embeddings}\label{appendix:dual_neural_networks}
To achieve better performance with fewer basis functions, we can also learn the basis functions $\widehat\psi(\cdot)$ and $\widehat\phi(\cdot)$ jointly with $\theta$ and $\eta$ by back-propagation. Specifically, denote the parameters in $\widehat\psi(\cdot) = [\widehat\psi_{\tilde w_1}(\cdot), \widehat\psi_{\tilde w_2}(\cdot), \ldots, \widehat\psi_{\tilde w_m}(\cdot)]^\top$ and $\widehat\phi(\cdot) = [\widehat\phi_{w_1}(\cdot), \widehat\phi_{w_2}(\cdot), \ldots, \widehat\phi_{w_m}(\cdot)]^\top $ explicitly as $\widetilde W = \sbr{\tilde w_i}_{i=1}^m$ and $W = \sbr{w_i}_{i=1}^m$, we also include $\tilde W$ and $W$ into optimization~\eq{eq:dual_random_fea}, which results
\begin{eqnarray}\label{eq:dual_neural_net}
\min_{\theta, \widetilde W}\max_{\eta, W} \widehat\Phi(\theta, \widetilde W, \eta, W):= \EE_{x, y}\EE_{z|x}\sbr{\theta^\top\widehat\psi_{\widetilde W}(z, x) \widehat\phi_W(x, y)^\top \eta -l_y^*\rbr{\eta^\top\widehat\phi_W(x, y)}} + \frac{\lambda_1}{2}\|\theta\|^2 - \frac{\lambda_2}{2}\|\eta\|^2.
\end{eqnarray}
Apply the proposed algorithm to~\eq{eq:dual_neural_net}, we obtain the update rule for all the parameters, $\{\theta, \eta, \widetilde W, W\}$, in $i$-th iteration,
\begin{equation*}
\begin{array}{rcl}
\theta_{i+1} &=& {(1 - \gamma_i\lambda_1)\theta_i-\gamma_i \eta_i^\top\widehat\phi_{W_i}(x_i, y_i)\widehat \psi_{\widetilde W_i}({z_i, x_i})},\\
\eta_{i+1} &=& (1 - \gamma_i\lambda_1)\eta_i + \gamma_i \sbr{\theta_i^\top\widehat \psi_{\widetilde W_i}(z_i, x_i) - \nabla \ell_{y_i}^*\rbr{\eta_i^\top\widehat\phi_{W_i}(x_i, y_i)}}\widehat \phi_{W_i}(x_i, y_i),\\
\widetilde W_{i+1} &=& \widetilde W_i - \gamma_i \eta_i^\top\widehat\phi_{W_i}(x_i, y_i)\theta_i^\top\nabla_{\widetilde W}\widehat \psi_{\widetilde W}({z_i, x_i}),\\
W_{i+1} &=& W_i + \gamma_i\sbr{\theta_i^\top \widehat \psi_{\widetilde W_i}({z_i, x_i}) - \nabla\ell_{y_i}^*\rbr{\eta_i^\top\widehat\phi_{W_i}(x_i, y_i)}}\eta_i^\top\nabla_W\widehat\phi_{W}(x_i, y_i).
\end{array}
\end{equation*}
Here we only demonstrate the back-propagation algorithm applies to one-layer basis functions, in fact, it can be extended to the deep basis functions, \ie, hierarchical composition functions, straight-forwardly if necessary. With such deep neural networks as function approximator in our algorithm, we achieve the dual neural networks embeddings.

\section{Experimental Details}\label{appendix:experiment_setup}

In all experiemnts, we conduct comparison on algorithms to optimize the objective with regularization on both primal and dual functions. Since the target is evaluating the performance of algorithms on the same problem, we fix the weights of the regularization term for the proposed algorithm and the competitors for fairness. The other paramters of models and algorithms, \eg, step size, mini-batch size, kernel parameters and so on, are set according to different tasks.

\subsection{Learning with Invariance}

\noindent{\bf Noisy in measurements.} We select the best $\eta \in \{0.1, 1, 10\}$ and $n_0\in \{1, 10, 100\}$. We use Gaussian kernel for both primal and dual function, whose bandwidth $\sigma$ are selected from $\{0.05, 0.1, 0.15, 0.2\}$. We set the batch size to be $50$. In testing phase, the observation is noisyless.

\noindent{\bf QuantumMachine.} We selected the stepsize parameters $\eta\in\{0.1, 0.5, 1\}$ and $n_0\in\{100, 1000\}$. We adopted Gaussian kernel whose bandwidth is selected by median trick with coeffcient in $\{0.1, 0.25, 0.5, 1\}$. The batch size is set to be $1000$. To illustrate the benefits of sample efficiency, we generated $10$ virtual samples in training phase and $20$ in testing phase. Notice that the number of virtual samples in this experiment is fewer than the one used in~\cite{DaiXieHe14}, therefore, the results are not directly comparable.

\subsection{Policy Evaluation}

We evaluated all the algorithms in terms of mean square Bellman error on the testing states. On each state $s$, the mean square Bellman error is estimated with 100 next states $s'$ samples. We set the number of the basis functions in GTD2 and RG to be $2^8$. To achieve the convergence property based the theorem~\cite{NemJudLanSha09}, we set stepsize to be be $\frac{\eta}{n_0 + \sqrt{t}}$ in the proposed algorithm and GTD2, $\frac{\eta}{n_0 + {t}}$ in Kernel MDP and RG.

\noindent{\bf Navigation.} The batch size is set to be $20$. $\{\eta, n_0\}\in \{0.1, 1, 10\}$. We adopted Gaussian kernel and select the best primal and dual kernel bandwidth in range $\{0.01, 0.05, 0.1, 0.15, 0.2\}$. The $\gamma$ in MDP is set to be $0.9$.

\noindent{\bf Cart-pole swing up.} The batch size is set to be $20$. The stepsize parameters are chosen in range $\{\eta, n_0\}\in \{0.05, 0.2, 1,10,50,100\}$. We adopted Gaussian kernel and the primal and dual kernel bandwidth are se lected by median trick with coeffcient in $\{0.5, 1,5,10\}$. The $\gamma$ is set to be $0.96$. The reward is $R(s)=\frac{1}{2}(s_1^2 + s_2^2 +s_3^2 +5(s_4-\pi)^2 )$ where the states $s_1,s_2,s_3,s_4$ are the cart position, cart velocity, pendulum velocity and pendulum angular position.

\noindent{\bf PUMA-560 manipulation.} The batch size is set to be $20$. The stepsize parameters are chosen in range $\{\eta, n_0\}\in \{0.05, 0.1, 5,10,100,500\}$. We adopted Gaussian kernel and the primal and dual kernel bandwidth are selected by median trick with coeffcient in $\{0.5, 1,5,10 \}$. The $\gamma$ is set to be $0.9$. The reward is 
$R(s) = \frac{1}{2}(\sum_{i=1}^4(s_i - \frac{\pi}{4})^2 + \sum_{i=5}^6(s_i + \frac{\pi}{4})^2 + \sum_{i=7}^{12}s_i^2)$
where  $s_{1},\ldots,s_{6}$ and $s_{7},\ldots,s_{12}$ are joint angles and velocities, respectively.

\section{Reinforcement Learning: Action Control}\label{appendix:control}

We can extend the same technique to control problem. Different from the policy evaluation in which the policy is provided, the control problem is trying to learn the optimal policy in terms of reward, \ie, the policy which can achieve the maximum reward. 

We define the action-value function $Q^\pi(s, a): \Scal \times \Acal \rightarrow \RR$, which evaluates the value of taking action $a$ in state $s$ with policy $\pi$ for future actions,  
$$
Q^\pi(s, a) = \EE\bigg[\sum_{t=0}^\infty \gamma^t R(s_{t+1}) \bigg| s_0 = s, A_0 = a, \pi\bigg].
$$
Based on the definition of the $Q^\pi$ function, we have the Bellman equation as
\begin{eqnarray}
Q^\pi(s, a) = R(s) + \gamma\EE_{s'\sim P(s' |s, a), a'\sim \pi(a|s')}[ Q^\pi(s', a')|s, a],
\end{eqnarray}
therefore, we obtain the optimization as 
\begin{eqnarray}\label{eq:sarsa_opt}
\min_Q L(Q) := \EE_{s\sim \mu(s), a\sim \pi(a|s)}\bigg[\bigg( \EE_{s', a'|s}[R(s) +\gamma  Q(s', a')] - Q(s, a) \bigg)^2\bigg] 
\end{eqnarray}
whose objective is the same as SARSA. Apply the dual kernel embedding, we obtain
\begin{eqnarray}\label{eq:sarsa_dual}
\min_Q\max_{u\in\Hcal}\EE_{s,a}\bigg[\bigg\langle \EE_{s', a'|s}[R(s) +\gamma  Q(s', a')] - Q(s, a), u(s, a) \bigg\rangle\bigg] - \frac{1}{2}\EE_{s, a}[u(s,a)^2],
\end{eqnarray}
which can be solved efficiently. Recall the definition of $Q$ function, our framework can be also extended to $\lambda$-return for the estimation of $R(s) +\gamma \max_{a'} Q(s', a')$, \ie, $G(\lambda) = (1-\lambda)\sum_{t=1}^\infty \lambda^{t-1}G^{(n)}$ where $G^{(n)} = \sum_{t=0}^n \gamma^t R(s_t)$.

With the proposed $Q$--function learning procedure, we can achieve the control target by \emph{policy iteration} in Algorithm~\ref{alg:policy_iteration}.
\begin{algorithm}[t]
\caption{Policy Iteration with $Q$-Dual Kernel Embedding}
  \begin{algorithmic}[1]\label{alg:policy_iteration}
    \STATE Initialize $Q(s, a)$ randomly
    \FOR{$i=1,\ldots, t$}
        \STATE update $\pi_i(a|s)$ as $\epsilon$-greedy policy from $Q_i(s, a)$
        \FOR{$j=1,\ldots, n$}
          \STATE sample $s\sim \mu(s)$, $a\sim \pi_i(a|s)$, $s'\sim p(s'|s, a)$ and $a'\sim \pi_i(a|s')$
          \STATE update $Q_i^j$ by applying Algorithm~\ref{alg:stochastic_composite_general} to~\eq{eq:sarsa_dual} 
        \ENDFOR
        \STATE $Q_{i+1} = Q_i^n$
    \ENDFOR
  \end{algorithmic}
\end{algorithm}
The convergence of this procedure can be proved by taking the approximate error from $Q$ function into account in the policy iteration convergence rate. \citet{PerPre02, Munos03} already provides a framework for the analysis. We omit the details here due to the major contribution of this paper is introducing a general sample efficient algorithm solving the optimization with nested expectation.

Besides the policy iteration procedure, we can also learn the optimal policy by solving Bellman optimal equation. We denote the optimal action-value function as $Q^*$,
$$
Q^*(s, a) = \max_\pi Q^\pi(s, a),\quad \forall s\in \Scal, a\in \Acal.
$$
Based on the definition of $Q^*$, we obtain the Bellman optimal equation for $Q^*$,
\begin{eqnarray*}
Q^*(s, a) = R(s) + \gamma\EE_{s'\sim P(s' |s, a)}[\max_{a'} Q^*(s', a')|s, a].
\end{eqnarray*}
Analogously, we can formulate the objective for $Q$-learning as 
\begin{eqnarray}\label{eq:qlearning_opt}
\min_Q L_q(Q) := \EE_{s\sim \mu(s), a\sim \pi_b(a|s)}\bigg[\bigg( \EE_{s'}[R(s) +\gamma \max_{a'} Q(s', a')] - Q(s, a) \bigg)^2\bigg] 
\end{eqnarray}
Obviously, for $Q$-learning, we need the extra attention that the objective is no longer convex due to the maximum operator, therefore, we cannot guarantee the global optimality.

\section{Policy Evaluation with Dynamics Kernel and Random Feature Conditional Embeddings}

In the policy evaluation experiment section, we compared the proposed algorithm with the MDPs embedding using kernels. \citet{GruLevBalPonetal12} introduces the RKHS embedding of transition dynamics in MDPs utilizing kernel conditional embedding~\cite{SonFukGre13}. With such nonparametric representation of the transition, the expectation can be computed easily and accurately by linear operatior, and thus, promoting the performance of policy evaluation and control task. For self-containing, we specify the details of MDPs embeddings with kernel and the random feature extension in this section. Moreover, to utilize data sequantially, we propose the algorithm for policy evaluation utilizing (functional) stochastic gradient with both kernel and random feature conditional embeddings, which is 
different from the value iteration algorithm with kernel embedded MDP in~\cite{GruLevBalPonetal12}.

\subsection{Dynamics RKHS Embeddings}
In MDPs, the transition dynamics, $p(s'|a, s)$, plays a vital. Many methods for solving MDPs requires the computation with respect to the dynamics, \ie, $\EE_{s'\sim p(s'|a, s)}[f(s')]$. By the kernel conditional embedding estimation~\cite{SonFukGre13}, such expectation can be estimated from data. Specifically, given data $\{(s_i, a_i, s_i')\}_{i=1}^m$, kernel $k:(\Xcal\times \Acal)\times (\Xcal\times \Acal)\rightarrow \RR$ and $\tilde k:\Xcal\times \Xcal\rightarrow \RR$, the expectation for $\forall f\in\tilde{\Hcal}$ can be approximated by $\langle \tau((s, a), \cdot), f \rangle$ with $ \tau((s, a), \cdot) \in \tilde{\Hcal}$ as 
\begin{eqnarray}\label{eq:kernel_conditional_embedding}
\tau_{(s, a)} = \alpha(s, a)^\top \widetilde K(S', \cdot),
\end{eqnarray}
where $\widetilde K(S', \cdot) = [\tilde k(s'_1, \cdot), \tilde k(s'_2, \cdot),\ldots, \tilde k(s'_m, \cdot)]^\top$, $\alpha(s, a) = (K+\lambda mI)^{-1} K((S, A), (s, a))$, $K((S, A), (s, a)) = [k((s_1, a_1), (s, a)), k((s_2, a_2), (s, a)), \ldots, k((s_m, a_m), (s, a))]^\top$ and $K = [k((s_i, a_i), (s_j, a_j))]_{i, j = 1}^m$, $\lambda$ is a regularization parameter. 

With the RKHS embedded transition dynamics, we can compute the expectation $\EE_{s'\sim p(s'|a, s)}[f(s')] \approx \langle \tau_{(s, a)}, f\rangle $. The theoretical property of such estimator is analyzed in~\cite{SonFukGre13,GruLevBalPatetal12}.

\subsection{Dynamics Random Feature Embeddings}

We can also leverage the random feature to approximate the kernel function in kernel conditional embedding, which will result the random feature embedding for memory efficiency. With the same notations, we denote the random features for $\tilde k$ and $k$ are $\widehat\psi_w(\cdot)$ and $\widehat\phi_w(\cdot)$ with respect to $\widetilde \PP(\omega)$ and $\PP(\omega)$ respectively, we can approximate the kernel conditional embedding~\eq{eq:kernel_conditional_embedding} by
\begin{eqnarray}
\widehat\tau_{(s, a)} = \widehat\phi(s, a)^\top(\widehat\Phi\widehat\Phi^\top + \lambda mI)^{-1}\widehat\Phi\widehat\Psi^\top,
\end{eqnarray}
where $\widehat\phi(s, a) = [\widehat\phi_{w_1}(s, a), \widehat\phi_{w_2}(s, a),\ldots, \widehat\phi_{w_p}(s, a)]^\top\in \RR^{p\times 1}$, $\widehat\psi(s) = [\widehat\psi_{\tilde w_1}(s), \widehat\psi_{\tilde w_2}(s),\ldots, \widehat\psi_{\tilde w_p}(s)]\in \RR^{p\times 1}$, $\widehat\Phi = [\widehat\phi(s_1, a_1), \widehat\phi(s_2, a_2), \ldots, \widehat\phi(s_m, a_m)]\in \RR^{p\times m}$ and $\widehat\Psi = [\widehat\psi(s_1'), \widehat\psi(s_2'), \ldots, \widehat\psi(s_m')]\in \RR^{p \times m}$. Then, with random feature represented $f(s') = \theta^\top \widehat\psi(s')$, the expectation $\EE_{s'\sim p(s'|a, s)}[f(s')]$ can be approximated by $\widehat\tau_{(s, a)}\theta$.

\subsection{Policy Evaluation with Dynamics Embeddings} 

As we introduced, with the kernel and random feature embedded transition dynamics, the expectation can be estimated with linear operator. Instead of plugging the embedded MDP into value iteration for policy evaluation in~\cite{GruLevBalPonetal12}, which is not suitable for dynamics random feature embeddings, we propose new algorithms by applying the (functional) stochastic gradient algorithm to optimize mean-square Bellman error with respect to $V(\cdot)$ in kernel representation or $\theta$ in random feature representation, we have 
\begin{eqnarray}\label{eq:SGD_gradient_MDP}
\nabla_{V(\cdot)} L &=& \rho(a|s)(V(s) - (R(s) + \gamma \langle \tau(s, a), V \rangle))(\tilde k(s, \cdot) - \gamma \EE_{s'|a, s}[\tilde k(s', \cdot)])\\
\nabla_\theta L &=& \rho(a|s)(\theta^\top \widehat\psi(s) - (R(s) + \gamma \widehat \tau(s, a)\theta))(\widehat\psi(s) - \gamma \widehat\tau(s, a)^\top)
\end{eqnarray}
where $\EE_{s'|a, s}[\tilde k(s', \cdot)]$ can be approximated by $\tau(a, s)$~\cite{SonFukGre13}. Therefore, plugging the gradient estimator into SGD results the following two algorithms for kernel/random feature embedded dynamics.

\begin{algorithm}[H]
  \caption{{SGD with MDP kernel embedding} }
    \text{\bf Input: $\pi(\cdot), \mu.$}\\[-4mm]
    \begin{algorithmic}[1]\label{alg:MDP_SGD_kernel}
      \STATE compute $\tau = (K + \lambda mI)^{-1}$
      \FOR{$i=1,\ldots, t$}
        \STATE Sample $s\sim\mu(s), a\sim\pi_b(a|s),s'\sim p(s'|s,a)$.
        \STATE $\bar v = K((S, A), (s, a))^\top\tau V_i(S')$.
        \STATE $\bar k(\cdot) = K((S, A), (s, a))^\top\tau \widetilde K(S', \cdot)$
        \STATE $V_{i+1} = (1 - \gamma_i\lambda_1)V_i - \gamma_i\rho(a|s)(V(s') - (R(s) + \gamma \bar v)(\tilde k(s, \cdot) - \gamma \bar k(\cdot))$.    
      \ENDFOR\\
    \end{algorithmic}
  \end{algorithm}

\begin{algorithm}[H]
  \caption{{SGD with MDP random feature embedding} }
    \text{\bf Input: $\pi(\cdot), \mu, \widehat\phi(\cdot), \widehat\psi(\cdot).$}\\[-4mm]
    \begin{algorithmic}[1]\label{alg:MDP_SGD_random_fea}
      \STATE compute $\tau = \rbr{\widehat\Phi\widehat\Phi^\top + \lambda mI}^{-1}\widehat\Phi\widehat\Psi^\top$
      \FOR{$i=1,\ldots, t$}
        \STATE Sample $s\sim\mu(s), a\sim\pi_b(a|s),s'\sim p(s'|s,a)$.
        \STATE $\theta_{i+1} = (1 - \gamma_i\lambda_1)\theta_i - \gamma_i\rho(a|s)(\theta_i^\top \widehat\psi(s') - (R(s) + \gamma \widehat\phi(s, a)^\top\tau\theta))(\widehat\psi(s') - \gamma\widehat\phi(s, a)^\top \tau)$.    
      \ENDFOR\\
    \end{algorithmic}
  \end{algorithm}

\end{appendix}

\end{document}